\documentclass[letterpaper, 10 pt, conference, twoside]{ieeeconf}

\IEEEoverridecommandlockouts  %
\let\labelindent\relax  %

\usepackage{booktabs}
\usepackage{diagbox}

\usepackage{graphicx} %
\usepackage{csquotes}

\makeatletter
\let\NAT@parse\undefined
\makeatother
\usepackage{xcolor,colortbl}
\definecolor{cvprblue}{rgb}{0.21,0.49,0.74}
\usepackage[colorlinks, breaklinks, citecolor=cvprblue]{hyperref}

\usepackage{mathtools}
\usepackage{amssymb}
\usepackage{svg}
\usepackage[hang,flushmargin]{footmisc}
\usepackage{cite}
\usepackage{microtype}
\usepackage{MnSymbol}  %
\usepackage{cuted}  %
\usepackage[export]{adjustbox}
\usepackage{pmboxdraw}
\usepackage{placeins}

\usepackage[inline]{enumitem}
\usepackage{multirow}
\usepackage{flushend}

\usepackage{siunitx}

\usepackage{subcaption}
\newcommand{\secref}[1]{Sec.~\ref{#1}}
\renewcommand{\eqref}[1]{Eq.~(\ref{#1})}
\newcommand{\figref}[1]{Fig.~\ref{#1}}
\newcommand{\tabref}[1]{Tab.~\ref{#1}}

\newcommand\subfiguresubref[1]{Subfig.~\subref{#1}}

\newcommand{\etal}{\emph{et al.}}

\newcommand{\hmm}{\mathcal{H}}

\newcommand{\mani}{\mathcal{M}}
\newcommand{\tans}{\mathcal{T}}
\newcommand{\mans}{\mathcal{S}}

\newdimen\figrasterwd
\figrasterwd\textwidth

\definecolor{svgGreen}{RGB}{0, 80, 0}

\newcommand{\rebuttal}[0]{}

\title{\LARGE \bf
The Art of Imitation: Learning Long-Horizon Manipulation Tasks from Few Demonstrations}
\author{Jan Ole von Hartz, Tim Welschehold, Abhinav Valada, and Joschka Boedecker
\thanks{Department of Computer Science, University of Freiburg, Germany}%
\thanks{This work was partially funded by Carl Zeiss Foundation with the ReScaLe project, the German Research Foundation (DFG): 417962828, and by the BrainLinks-BrainTools center of the University of Freiburg.}
}
\date{March 2024}

\begin{document}
\maketitle

\begin{abstract}
    Task Parametrized Gaussian Mixture Models (TP-GMM) are a sample-efficient method for learning object-centric robot manipulation tasks.
    However, there are several open challenges to applying TP-GMMs in the wild.
    In this work, we tackle three crucial challenges synergistically.
    First, end-effector velocities are non-Euclidean and thus hard to model using standard GMMs.
    We thus propose to factorize the robot's end-effector velocity into its direction and magnitude, and model them using Riemannian GMMs.
    Second, we leverage the factorized velocities to segment and sequence skills from complex demonstration trajectories.
    Through the segmentation, we further align skill trajectories and hence leverage time as a powerful inductive bias.
    Third, we present a method to automatically detect relevant task parameters \emph{per} skill from visual observations.
    Our approach enables learning complex manipulation tasks from just five demonstrations while using only RGB-D observations.
    Extensive experimental evaluations on RLBench demonstrate that our approach achieves state-of-the-art performance with 20-fold improved sample efficiency.
    Our policies generalize across different environments, object instances, and object positions, while the learned skills are reusable.\looseness=-1 
\end{abstract}

\section{Introduction}
Learning robot manipulation tasks from a handful of demonstrations is a challenging problem~\cite{celemin2022interactive, chisari2022correct}.
This problem is exacerbated when learning only from visual observations and aiming to generalize across task instances.
Deep learning approaches can be parameterized flexibly and generalize well, but usually require hundreds of demonstrations.
Leveraging compact observation spaces, such as scene keypoints, alleviates the problem, but does not resolve it entirely~\cite{vonhartz2023treachery}. %
In contrast, Gaussian Mixture Models (GMMs)~\cite{calinon2006learning} can be learned from only a handful of demonstrations.
Various extensions to the GMM framework exist to enable generalization across task instances.
For example, Task-Parameterized GMMs (TP-GMM)~\cite{calinon2013improving} generalize by modeling the set of demonstrations from the perspective of multiple relevant coordinate frames, called \emph{task-parameters}. %

\begin{figure}
    \centering
    \includegraphics[width=\columnwidth]{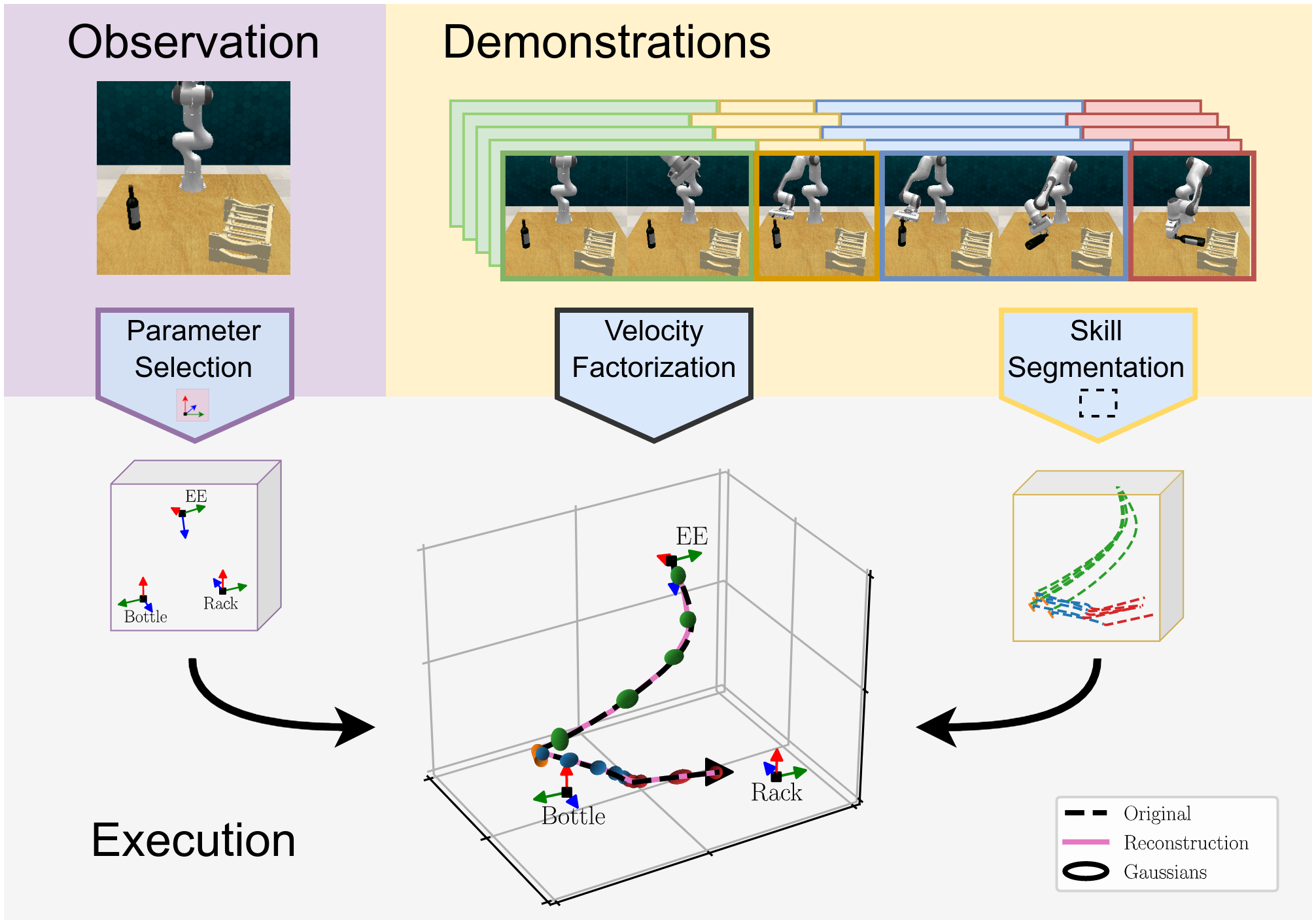}
    \caption{\textit{TAPAS-GMM}: Task Auto-Parameterized And Skill Segmented GMM learns task-parameterized manipulation policies from only a handful of complex task demonstrations.
    First, we segment the full task demonstrations into the involved skills.
    For each segment, we then automatically select the relevant task parameters and learn a Riemannian Task-Parameterized Hidden Markov Model (TP-HMM).
    The skill models can be cascaded and reused flexibly.
    To enable modeling of the robot's end-effector velocity, we further leverage a novel action factorization and Riemannian geometry.}
    \label{fig:overview}
   \vspace{-0.3cm}
\end{figure}

Current methods based on TP-GMMs exhibit a set of serious limitations.
First, velocity information is often important for effectively modeling robotic tasks.
However, velocities are best expressed by their direction and magnitude, and are poorly represented using  Euclidean approaches.
As GMMs typically assume Euclidean data, velocity information is difficult to model using TP-GMMs~\cite{sun2023damm}.
Second, GMMs suffer from the \emph{curse of dimensionality},
i.e.\ they scale poorly to both high-dimensional data and long trajectories.
Both are crucial drawbacks in a robotics context.
On the one hand, the dimensionality of the TP-GMM scales linearly with the number of task parameters.
On the other hand, even seemingly simple tasks such as \emph{pick and place} can be too long and complex to fit.
Third, tasks often consist of multiple skills such as \emph{pick up the bottle}.
These skills might not be temporally aligned across demonstrations, rendering the data distribution even more complex.
Fourth, when learning from visual observations, generating a set of candidate task parameters is an open problem~\cite{calinon2016tutorial, li2023task}.
In practice, object tracking systems are used~\cite{zeestraten2018programming}, but this hardware dependency severely limits the model's applicability.
Finally, the task-parameter selection is crucial for the resulting model quality.
However, existing approaches for selecting among a set of candidates also suffer from the curse of dimensionality~\cite{alizadeh2014learning}.

We propose Task Auto-Parameterized And Skill Segmented GMM \textit{TAPAS-GMM}, illustrated in \figref{fig:overview}.
Our method addresses this set of challenges synergistically.
First, we propose to refactorize the robot's velocity into direction and magnitude to better capture the underlying data distribution.
We then model both using Riemannian GMMs~\cite{zeestraten2018programming}, which are not constrained by the Euclidean assumptions. %
Second, we segment the task demonstrations into skills using the velocity information.
The segmentation both aligns the trajectories temporally and enables learning skill models with shorter trajectories and lower data dimensionality.
Additionally, the temporal alignment enables us to use time as a powerful inductive bias for the initialization of our model.
If the trajectories are aligned, then simple binning of the trajectories over the time dimension yields an excellent initialization.
Third, we use a pretrained visual encoder network~\cite{amir2021deep} to generate object keypoints that generalize across object instances and task environments~\cite{vonhartz2023treachery}.
This way, we generate a set of \emph{candidate} task parameters from RGB-D observations.
Finally, for selecting the task-relevant parameters from these candidates, we offer an approach that scales well to large numbers of candidates.
Our proposed solutions have synergistic effects.
For example, combining our parameter selection and skill segmentation further reduces the data dimensionality, as a given task parameter is often only relevant for \emph{some} skill and not the whole long-horizon task. 

We evaluate our method on a diverse set of manipulation tasks, both in RLBench~\cite{james2019rlbench} and on a real robot, including articulated objects, as well as high precision and long-horizon tasks.
We experimentally disentangle policy learning and representation learning, as well as ablating the major components of our approach.
\textit{TAPAS-GMM} learns long-horizon manipulation policies from as little as five demonstrations. 
We show generalization across object positions, object instances,  visual clutter, and task environments.
Finally, we demonstrate how the learned skills can be recombined in novel ways.

In summary, our main contributions are
\begin{enumerate}
    \item a novel action factorization to effectively model end-effector velocities using Task-Parameterized Gaussian Mixture Models.
    \item a new velocity-based approach to segmenting complex demonstration trajectories into skills.
    \item an approach that efficiently extracts and selects task parameters per skill from visual observations.
    \item rigorous evaluation of our approach's ability to overcome the problems outlined above and its efficacy for policy learning, both in simulation and on a real robot.
    \item We make the code and models publicly available at \url{http://tapas-gmm.cs.uni-freiburg.de}.
\end{enumerate}
The supplementary material is appended at the end of this paper.
\section{Related Work}
{\parskip=3pt
\noindent\textit{Parameterized Motions:} GMMs~\cite{calinon2006learning} learn robotic manipulation tasks from only a handful of demonstrations.
Popular alternatives are Dynamic Motion Primitives (DMP)~\cite{ijspeert2013dynamical} and Probabilistic Motion Primitives (ProMP)~\cite{paraschos2013probabilistic}.
However, all three methods require additional techniques to adapt to novel task instances. %
Calinon~\etal{}~\cite{calinon2013improving} identify three classes of adaptation methods: model database, parametric, and multi-stream, to which we amend two-stage and point-based approaches.
Model database approaches~\cite{matsubara2011learning} learn one model per demonstration, that they retrieve and combine based on the associated parameters.
In contrast, parametric models~\cite{pervez2018learning} learn a single, parameterized model.
Two-stage approaches first learn a model encoding a primitive motion that they adapt in a second stage, e.g., \ using reinforcement learning~\cite{nematollahi2022robot} or Wasserstein gradients~\cite{ziesche2024wasserstein}.
However, this requires a second online learning phase.
Whereas, multi-stream approaches~\cite{calinon2007learning} model the same set of demonstrations from the perspective of multiple coordinate frames, called \emph{task-parameters}.
They generalize via the transformation and combination of these local models.
Task-parameterized Gaussian Mixture Models (TP-GMM)~\cite{calinon2013improving, zeestraten2018programming, rozo2020learning} are multi-stream methods, leveraging a single expectation-maximization procedure.
Finally, point-based approaches~\cite{li2023task, huang2019kernelized, heppert2024ditto} warp the underlying model to explicit endpoints and waypoints.}

TP-GMMs estimate the relevance of a task parameter for a given observation from the variance observed in the task demonstrations.
Consequently, they generalize poorly when given fewer than a handful of demonstrations~\cite{li2023task}.
In contrast, directly editing waypoints~\cite{li2023task, huang2019kernelized} is feasible for just a single demonstration.
However, task parameters can be extracted from visual observations~\cite{chisari2023centergrasp}, while defining waypoints may require significant engineering effort.
Moreover, by using multiple demonstrations, TP-GMMs approximate the full trajectory distribution suited to solve a given task, instead of assuming the optimality of a single demonstration.
Most importantly, TP-GMMs can be extended to model rotations~\cite{zeestraten2018programming}, which are required to effectively solve many robotic manipulation tasks.
Whereas, both Elastic-DS~\cite{li2023task} and KMP~\cite{huang2019kernelized} are only defined for Euclidean spaces.

{\parskip=3pt
\noindent\textit{Parameter Selection:} Task-parameters are usually \enquote{assumed to be known or given during demonstrations}~\cite{li2023task}.
This requires \enquote{the experimenter to provide an initial set of [potentially relevant candidate] frames}~\cite{calinon2016tutorial}.
Given a set of candidate frames, their respective relevance for a given demonstration can be estimated~\cite{ alizadeh2016identifying}.
However, current approaches fit a joint model to all candidate frames~\cite{alizadeh2016identifying}.
Hence, they suffer from the \emph{curse of dimensionality}, which our method avoids.
Furthermore, automatically generating an initial set of candidate frames is an open problem~\cite{calinon2016tutorial, li2023task}.
In practice, infrared motion tracking devices are manually attached to relevant objects~\cite{zeestraten2018programming}, limiting the practical applicability of these methods.
In contrast, we propose to automatically identify candidate task parameters from visual observations and to select those relevant to the task at hand.
This enables instance-level generalization across objects and applications in novel and changing environments.}

{\parskip=3pt
\noindent\textit{Sample-Efficient Long-Horizon Manipulation:}
GMMs are typically used to model full  trajectories~\cite{nematollahi2022robot}, with some work focusing on solving long-horizon tasks by sequencing learned GMMs in a bottom-up manner~\cite{rozo2020learning}.
In contrast, we focus on directly learning long-horizon tasks by segmenting and sequencing skills \emph{top-down} from complex task demonstrations.
Some deep learning based approaches~\cite{shridhar2023perceiver, goyal2023rvt} simplify long-horizon problems by only predicting a set of key poses and using motion planning in between.
However, by throwing away the trajectory information between key poses, they cannot produce constraint motions such as opening a cabinet.
Moreover, they can only predict key poses in the demonstrated order, whereas our approach further produces a set of reusable and re-combinable skills.}
\section{Background}
In this section, we summarize foundational work on Gaussian Mixture Models for robotic manipulation. %
We provide additional details including full equations in \secref{sec:gmreqs} of the supplementary material.

\begin{figure}[tb]
    \centering
    \includegraphics[width=0.31\columnwidth,valign=t]{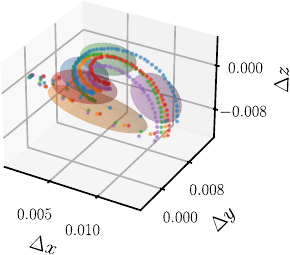}\hfil
    \includegraphics[width=0.31\columnwidth,valign=t]{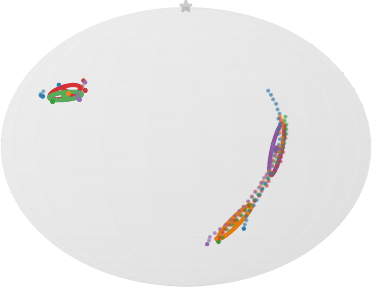}\hfil
    \includegraphics[width=0.31\columnwidth,valign=t]{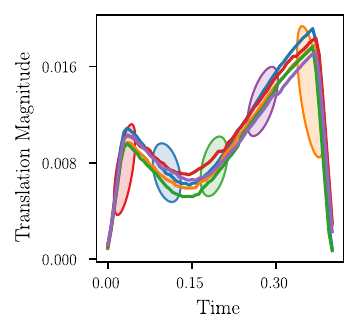}
    \caption{Velocity trajectories from the first skill in \texttt{StackWine}.
    \textit{Left:} The velocities are difficult to cluster in Euclidean space.
    \textit{Middle:} \(\mans^2\) models the movement direction.
    \textit{Right:} The associated action magnitudes.
    }\label{fig:distcomp}
   \vspace{-0.3cm}
\end{figure}

\subsection{Gaussian Mixture Model}
A Gaussian Mixture Model (GMM) with  \rebuttal{the Hidden Markov Model (HMM) extension~\cite{calinon2010hmm} and} \(K\) Gaussian components is defined as \(\mathcal{G}=\{\pi_k, \rebuttal{\{a_{kl}\}_{l=1}^K}, \boldsymbol\mu_k, \boldsymbol\Sigma_k\}_{k=1}^K\).
\(\boldsymbol\mu_k\) and \(\boldsymbol\Sigma_k\) denote the \(k\)-th component's mean and covariance matrix and \(\pi_k\) denotes its prior probability.
\rebuttal{\(a_{kl}\) is the transition probability from state \(k\) to state \(l\).}
In the context of Learning from Demonstrations, GMMs are employed to estimate the joint distribution of a disjoint set of input variables \(\mathcal{I}\) and output variables \(\mathcal{O}\), from a set of task demonstrations \(\mathcal{D}\) as
\begin{equation}\label{eq:gmm}
    p_\mathcal{G} (\mathcal{D}) = p(\mathcal{I}, \mathcal{O} \mid \mathcal{G}) = \sum_{k=1}^K \pi_k \cdot \mathcal{N}(\mathcal{I}, \mathcal{O}\mid \boldsymbol\mu_k, \boldsymbol\Sigma_k).
\end{equation}
We define \(\mathcal{D}=\{\{\boldsymbol z_t^n\}_{t=1}^T\}_{n=1}^N\), where \(\boldsymbol z\in\mathcal{I}\cup\mathcal{O}\), while \(t\) indexes time and \(n\) the demonstration.
We use Expectation Maximization (EM) to maximize the probability of \(\mathcal{D}\) under \(\mathcal{G}\). 
For inference, we employ Gaussian Mixture Regression (GMR)~\cite{cohn1996active} to estimate the conditional density %
\begin{equation}\label{eq:gmr}
    p_\mathcal{G} (\boldsymbol o\mid\boldsymbol i) = \sum_{k=1}^K \rebuttal{\pi_k^{\boldsymbol i}} \cdot\mathcal{N}(\boldsymbol o\mid \boldsymbol\mu_k ^{o\mid i}, \boldsymbol\Sigma_k^{o\mid i}).
\end{equation}

Next, we discuss the choice of input and output variables.

\subsection{Time-Driven vs.\ State-Driven Models}\label{sec:tvsx}
There are at least two ways to model robotic GMM policies: time-driven and state-driven.
Let \(\boldsymbol\xi\) denote the robot's end-effector pose.
Time-driven GMMs model the robot's state conditional on time: \(p(\boldsymbol\xi\mid t)\), i.e.\ \(\mathcal{I} = \{t\}\) and \(\mathcal{O}=\{\boldsymbol\xi\}\).
As they do not model the robot's velocity, we post-process the predictions to ensure compliance with the robot's velocity limits, e.g.\ using TOPPRA~\cite{pham2018new}.

State-driven GMMs model the robot's velocity given its current state: \(p(\dot{\boldsymbol\xi}\mid \boldsymbol\xi)\), i.e.\ \(\mathcal{I} = \{\boldsymbol\xi\}\) and \(\mathcal{O}=\{\dot{\boldsymbol\xi}\}\), which also captures the robot's velocity limits.
However, velocities are harder to model because they are non-Euclidean.
Specifically, velocity trajectories are usually less smooth and have more multimodalities than state trajectories.
Moreover, velocity errors can aggregate over time.
Additionally, deployment on a robot can require time-independence of the policy, e.g.\ to effectively deal with disturbances.
We evaluate these aspects experimentally in \secref{sec:state_res}.

\begin{figure}[tb]
    \centering
    \includegraphics[width=0.4\columnwidth]{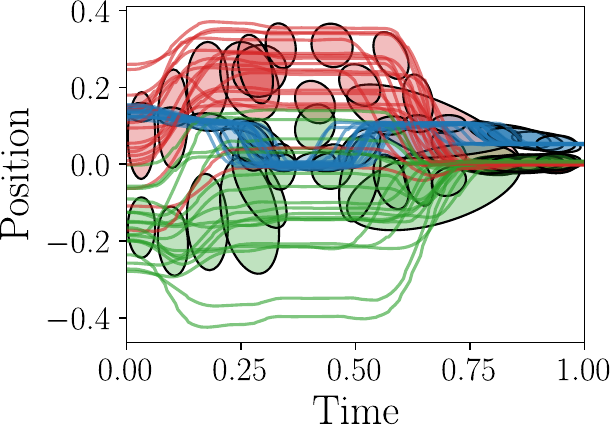}\hfil
    \includegraphics[width=0.4\columnwidth]{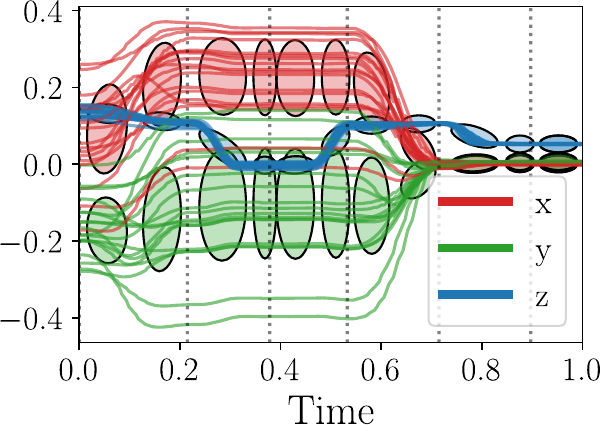}
    \caption{\textit{Left:} GMM on unaligned demos. \textit{Right:} GMM on aligned demos.
    Dotted lines indicate segment borders. 
    }\label{fig:tempalign}
   \vspace{-0.3cm}
\end{figure}

\subsection{Task-Parameterization}\label{sec:tpgmm}
Task-Parameterized GMMs~\cite{calinon2013improving} model the movement from the perspective of multiple coordinate frames, called \emph{task-parameters}, in one joint GMM~\cite{zeestraten2018programming}.
These frames are attached to task-relevant objects but may also include the robot's frame and a world frame.
Let \(\boldsymbol x (f)\) denote the end-effector \emph{position} in the frame of task-parameter \(f\).
We discuss the extension to full \emph{poses} \(\boldsymbol\xi(f)\) in \secref{sec:riemann}.
For the time-driven model and \(F\) task-parameters, we model \(\mathcal{I}=\{t\} \text{ and } \mathcal{O} = \{\boldsymbol x (f)\}_{f=1}^F\).
In contrast, for the state-driven model, we define
\(\mathcal{I} = \{\boldsymbol x (f)\}_{f=1}^F \text{ and } \mathcal{O} = \{\dot{\boldsymbol x} (f)\}_{f=1}^F\).
Akin to the vanilla GMM, we fit all parameters jointly using EM.
Next, we compute the per-frame \emph{marginal} model \rebuttal{\(p_f(\mathcal{I}, \mathcal{O} \mid\mathcal{G})= \left\{\pi_k^f, \boldsymbol\mu_k^f, \boldsymbol\Sigma_k^f\right\}_{k=1}^K\)} for each frame \(1\le f\le F\).
To adapt to a given set of task parameters \(\Theta=\{\boldsymbol A_f, \boldsymbol b_f\}_{f=1}^F\), we transform the per-frame marginals to the world frame \rebuttal{as}
\begin{equation}\label{eq:frametrans}
\begin{split}
    p_f(\mathcal{I}, \mathcal{O} \mid\mathcal{G},\Theta) &= \left\{\pi_k^f, \hat{\boldsymbol\mu}^f_k, \hat{\boldsymbol\Sigma}^f_k\right\}_{k=1}^K,\\
    \hat{\boldsymbol\mu}^f_k &= \boldsymbol A_f\boldsymbol\mu_{k}^f +\boldsymbol b_f,\\
    \hat{\boldsymbol\Sigma}^f_k &= \boldsymbol A_f \boldsymbol\Sigma_k^f\boldsymbol A_f^T.
\end{split}
\end{equation}
Finally, we compose the joint model as
\begin{equation}\label{eq:join}
    p_\mathcal{G} (\mathcal{I}, \mathcal{O}\mid\Theta) =\prod_{f=1}^F p_f(\mathcal{I}, \mathcal{O} \mid\mathcal{G},\Theta).
\end{equation}
We then perform GMR on the joint model as defined in \eqref{eq:gmr}.
For an excellent, more detailed overview of TP-GMMs, please refer to \cite{calinon2016tutorial}.

\subsection{Riemannian Manifolds}\label{sec:riemann}
We model the robot's end-effector pose \(\boldsymbol\xi=\begin{bmatrix}\boldsymbol x & \boldsymbol \theta\end{bmatrix}^T\) on the Riemmanian Manifold \(\mani_{\text{pose}}=\mathbb{R}^3\times\mathcal{S}^3\).
Whereas GMMs typically assume Euclidean data, the naive Euclidean mean of a set of quaternions is generally not a valid quaternion.
Each Riemannian manifold \(\mani\) is differentiable, i.e., at each point \(\mathbf{x}\), it has a tangent space \(\tans_\mathbf{x}\mani\), that is a real vector space.
\rebuttal{
Zeestraten~\cite{zeestraten2018programming} formulates all previously Euclidean operations in EM and GMR, including the quaternion center of mass, as non-linear Maximum Likelihood Estimation problems on the tangent space.
We provide additional details in \secref{sec:mandetails} of the supplementary material and Zeestraten~\cite{zeestraten2018programming} gives an excellent introduction into Riemannian Manifolds.
Note that parallel transport is used to implement the frame transformations.
We found this to sometimes introduce unwanted correlations into the model's covariance, leading to faulty predictions.}
We discuss the problem and our proposed solution in detail in \secref{sec:covtrans} of the supplementary material.
\section{Technical Approach}
We aim to learn robot manipulation policies that generalize to unseen task instances from just five demonstrations.
To generalize across object instances and be independent of object-tracking hardware, the task parameters should be extracted from RGB-D observations.
For fast and easy data collection, complex tasks should automatically be segmented into their atomic skills.
The method should not require one collection phase per atomic skill and should be able to construct novel sequences that were not demonstrated.

\begin{figure}[tb]
    \centering
    \includegraphics[width=\columnwidth]{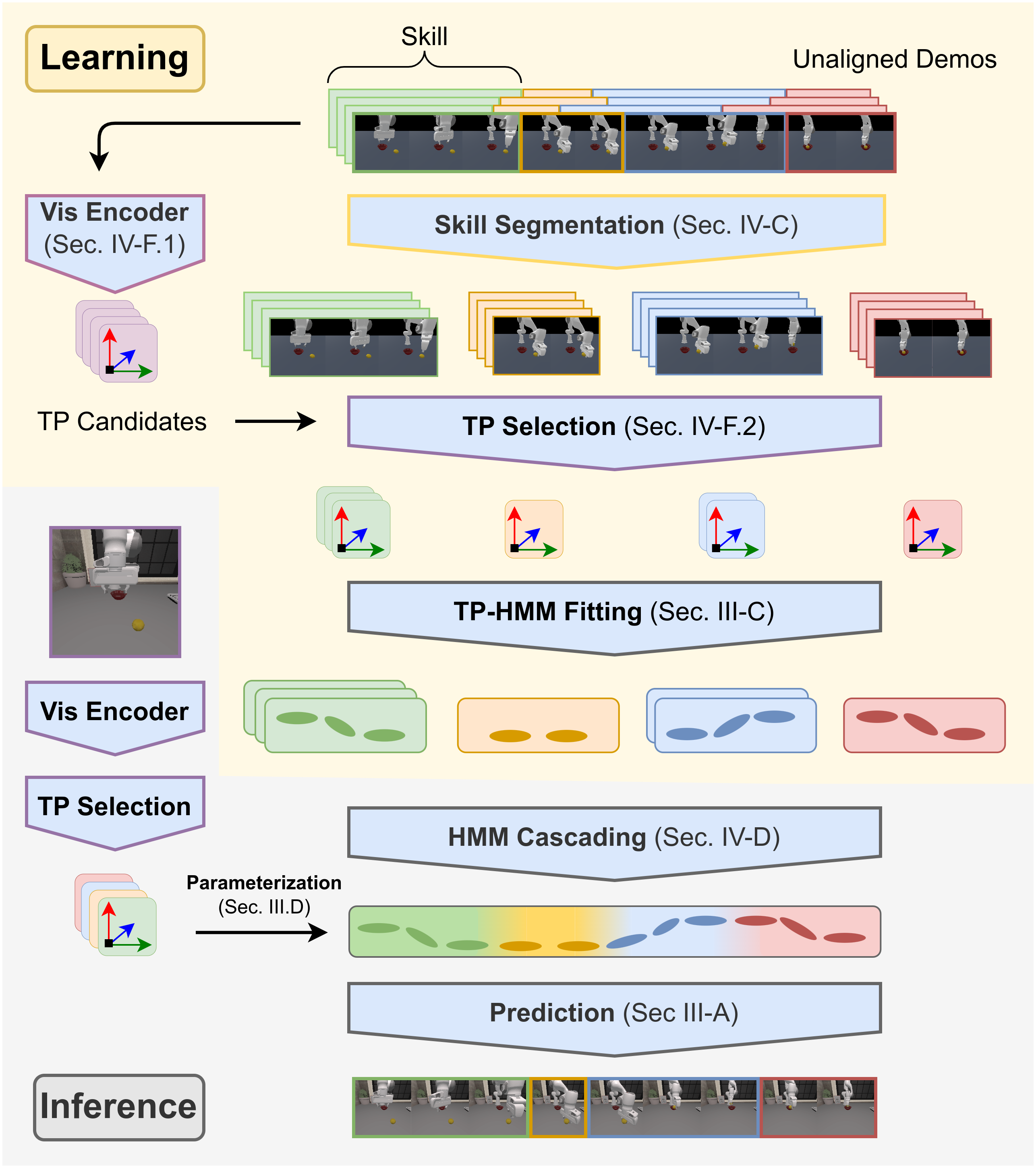}
    \caption{Overview of our approach.
    \textit{Learning:} First, we segment a set of complex task demonstrations with unaligned skills.
    Next, we generate a set of candidate task parameters from visual observations and select the relevant parameters for each segment.
    Finally, we fit one Task-Parameterized Hidden Markov Model (TP-HMM) per segment.
    \textit{Inference:} To make a prediction for new visual observations, we again extract the set of task parameters and select the task parameters determined during the learning phase.
    We then cascade the segment TP-HMMs.}
    \label{fig:flow}
   \vspace{-0.3cm}
\end{figure}

We present \textit{TAPAS-GMM}: Task-Parameterized and Skill Segmented GMM.
Our approach is based on Riemannian Task-Parameterized Hidden Markov Models.
To effectively model velocity data, we propose a novel factorization.
We then segment skills from complex demonstrations to temporally align the demonstrations and to learn skill models that we can sequence and reuse later.
Finally, we propose an approach to automatic task parameterization and combine these techniques in a synergistic manner.
\figref{fig:flow} summarizes our approach.
We now introduce the components in turn.

\begin{figure}[tb]
    \centering
    \includegraphics[width=\columnwidth, trim={0, 0.25cm 0 0}, clip]{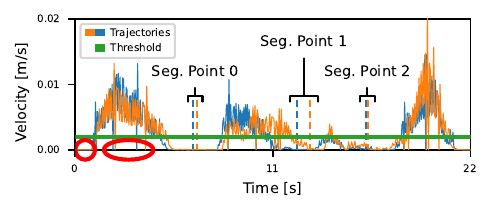}%
   \caption{\rebuttal{Velocity-based skill segmentation on two noisy robot trajectories for a pouring task.
   Red ellipses indicate candidate segmentation points that were filtered out: the first set is too close to the start of the trajectory before the robot begins moving, while the second set results from noise in the trajectory.
   The final segmentation points are the centers of extended sub-threshold segments.}
    }\label{fig:skillseg}
   \vspace{-0.3cm}
\end{figure}

\begin{figure*}[tb]
    \centering
    \includegraphics[width=0.075\textwidth,valign=t]{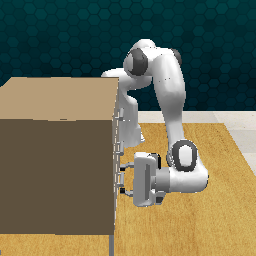}\hfil
    \includegraphics[width=0.075\textwidth,valign=t]{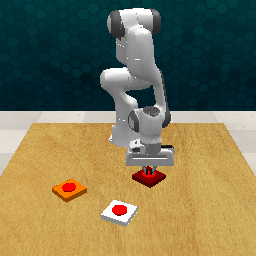}\hfil
    \includegraphics[width=0.075\textwidth,valign=t]{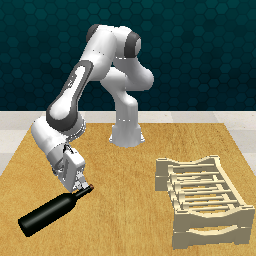}\hfil
    \includegraphics[width=0.075\textwidth,valign=t]{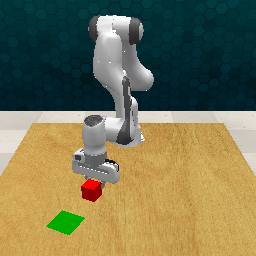}\hfil
    \includegraphics[width=0.075\textwidth,valign=t]{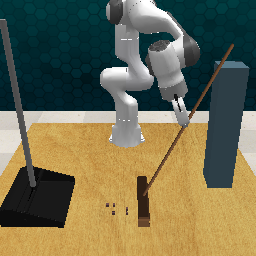}\hfil
    \includegraphics[width=0.075\textwidth,valign=t]{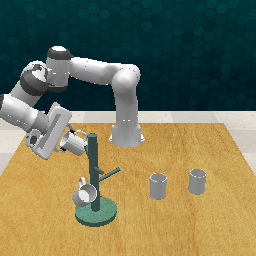}\hfil
    \includegraphics[width=0.075\textwidth,valign=t]{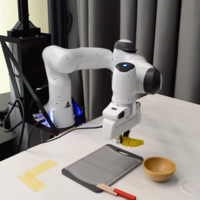}\hfil
    \includegraphics[width=0.075\textwidth,valign=t]{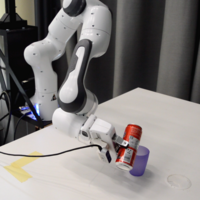}\hfil
    \includegraphics[width=0.075\textwidth,valign=t]{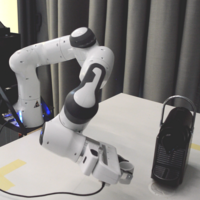}\hfil
    \includegraphics[width=0.075\textwidth,valign=t]{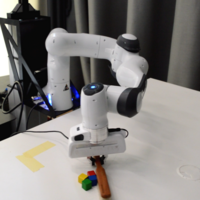}\hfil
    \includegraphics[width=0.075\textwidth,valign=t]{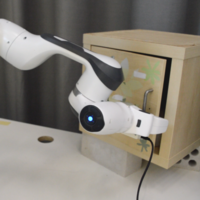}\hfil
    \includegraphics[width=0.075\textwidth,valign=t]{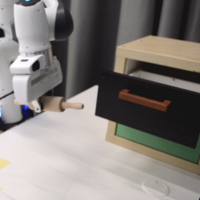}\hfil
    \includegraphics[width=0.075\textwidth,valign=t]{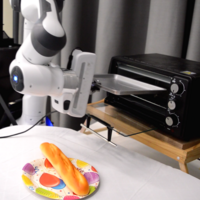}
    \caption{
    \textit{Left:} The RLBench tasks \texttt{OpenDrawer}, \texttt{PushButtons}, \texttt{StackWine}, \texttt{SlideBlock}, \texttt{SweepToDustpan}, and \texttt{PlaceCups}. \textit{Right:} The real-world tasks \texttt{PickAndPlace}, \texttt{PourDrink}, \texttt{MakeCoffee}, \texttt{SweepBlocks}, \texttt{OpenCabinet}, \texttt{StoreInDrawer}, and \texttt{BakeBread}.}\label{fig:tasks}
   \vspace{-0.3cm}
\end{figure*}

\subsection{Action Factorization}\label{sec:xfact}
In \secref{sec:tvsx}, we discussed that solving manipulation tasks can require modeling the end-effector's velocity \(\dot{\boldsymbol\xi}\) and not only its pose \(\boldsymbol\xi\), e.g.\ if time-independence is required.
\figref{fig:distcomp} illustrates that naively modeling \(\dot{\boldsymbol\xi}\) on the manifold \(\mani_{\text{pose}}=\mathbb{R}^3\times\mathcal{S}^3\) (introduced in \secref{sec:riemann}) is difficult because velocities are not Euclidean.
Instead, they are vector quantities defined by a direction and magnitude.
We therefore propose to factorize the end-effector velocity \(\dot{\boldsymbol\xi}\) as
\begin{equation}
    \dot{\boldsymbol\xi}=\begin{bmatrix}\hat{\dot{\boldsymbol x}} & \hat{\dot{\boldsymbol\theta}} & \left\lVert\dot{\boldsymbol x}\right\rVert & \left\lVert\dot{\boldsymbol\theta}\right\rVert\end{bmatrix}^T.
\end{equation}
The translation direction \(\hat{\dot{\boldsymbol x}}\) is the normalized translation vector, whereas \(\hat{\dot{\boldsymbol\theta}}\) is the rotation \emph{axis}.
Both lie on \(\mans^2\).
The translation magnitude \(\left\lVert\dot{\boldsymbol x}\right\rVert\in\mathbb{R}\) is the norm of the translation vector and \(\left\lVert\dot{\boldsymbol\theta}\right\rVert\in\mans^1\) is the rotation angle.
The combined manifold is \(\mathcal{M}_{\text{vel}} = \mathcal{S}^2\times\mathcal{S}^2\times\mathbb{R}\times\mathcal{S}^1\).
For a set \(F\) of task parameters, the full manifold of our state-driven TP-GMM is given by
\begin{equation}\label{eq:xdxmani}
    \mani=(\mani_{\text{pose}}\times\mans^2\times\mans^2)^F \times\mathbb{R}\times\mans^1.
\end{equation}

Our insight is three-fold.
First, in the task-parameterized setting, only the direction of the velocities varies between different fixed coordinate frames, not their magnitude.
Thus, we reduce the dimensionality of the problem by modeling the magnitudes globally.
\rebuttal{Second, end-effector velocities are not Euclidean due to the curvature of the trajectories and the motion constraints and smooth acceleration of the robot arm.
Consequently, the velocity can be modeled more effectively when factorized, as illustrated in \figref{fig:distcomp}.
We validate this aspect empirically in \secref{sec:state_res}.}
To model the factorized velocities, we leverage the Riemannian toolbox introduced in \secref{sec:riemann}.
Third, we leverage the isolated velocities to automatically segment complex task trajectories into the involved skills.
We found this to drastically improve the quality and transferability of the learned model and discuss it in detail in \secref{sec:trajseg}.
We provide additional details on how to implement the frame transformations for \(\mans^2\) in \secref{sec:mandetails} of the supplementary material.

\subsection{Gripper Action}
Effectively solving manipulation tasks also requires modeling the gripper action.
To this end, we add another global action dimension \(\mathbb{R}\), modeling the gripper width, to \eqref{eq:xdxmani}.

\subsection{Skill Segmentation}\label{sec:trajseg}
Robot manipulation tasks often involve a set of skills.
The pick and place task in \figref{fig:flow} consists of four skills.
\begin{enumerate*}
    \item Align the gripper with the first object.
    \item Grasp it.
    \item Align the first object with the second object.
    \item Place the first object.
\end{enumerate*}
All skills have distinct dynamics, so it is difficult to model them jointly in a single \emph{state}-driven GMM.
Furthermore, the skills have different relative durations across task instances, leading to poor temporal alignment across demonstrations.
\figref{fig:tempalign} illustrates that, as a consequence, they are difficult to model with a single \emph{time}-driven GMM.

Therefore, we propose to automatically segment the full demonstration trajectories into a sequence of skill demonstrations and to learn one model per skill.
This allows us to temporally align the skills and to learn local dynamics, specific to the skills.
The skill model can be time-driven or state-driven.
Furthermore, it allows us to reuse the learned skills by sequencing them in novel ways.
Lastly, in \secref{sec:init} and \secref{sec:framesel} we discuss its synergistic effects with our initialization strategy and our automatic task-parameterization, respectively.
To segment the trajectories, we threshold the action magnitudes introduced in \secref{sec:xfact}.
\rebuttal{We cluster observations with near-zero action magnitudes into time-based segments, yielding a set of candidate segmentation points.
We then filter out candidates that are too close to the beginning or end of each trajectory, as well as segments below a minimum length threshold.
As shown in \figref{fig:skillseg}, short segments, often only a single time step long, are typically the result of noise in the trajectory, whereas valid segmentation points correspond to extended periods of sub-threshold velocity.
We then use the temporal centers of the remaining segments as segmentation points, assuming that skills are presented in the same order across demonstrations.
Afterward, we align the skill demonstrations by resampling them to match their average duration.
Finally,} we fit one TP-HMM per skill.

\subsection{Skill Sequencing and Skill Reuse}\label{sec:sequence}
We execute a sequence of skills by cascading the associated HMMs.
To cascade two HMMs \(\hmm_1,\hmm_2\), Rozo \etal{} astutely propose to calculate the transition probability \(a_{ij}\) between state \(i\in\hmm_1\) and state \(j\in\hmm_2\) using the KL-divergence between the associated Gaussians\cite{rozo2020learning} as
\begin{equation}
    a_{ij} \propto \exp\left(- D_{KL}(\mathcal{N}(\boldsymbol\mu_i, \boldsymbol\Sigma_i)\parallel \mathcal{N}(\boldsymbol\mu_j, \boldsymbol\Sigma_j))\right).
\end{equation}
In case \(\hmm_1\) and \(\hmm_2\) contain different task parameters, we marginalize the Gaussians to their set of common task parameters first.
Afterward, we normalize the outgoing probabilities of all states.
Computing the KL-divergence can be difficult for Riemannian Gaussians, so we estimate it efficiently using a Monte-Carlo sampling scheme~\cite{hershey2007approximating}.
We employ this procedure iteratively to longer sequences of HMMs.
Note that this skill sequencing procedure does not require the skills to be sequenced in the demonstrated order.
Instead, it allows us to solve new tasks by reordering and sequencing learned skills.
Note further that time-based GMM skills can easily be reversed, either by manipulating the mean and covariance matrices or by reversing the time signal.
In combination, this allows us to create a rich set of new task policies, which we demonstrate in \secref{sec:real_world_gen}.

\begin{table*}
        \caption{Success rates of the learned policies in simulated tasks. Results for 3D Diffuser Actor were reported by the authors.}\label{tab:success_rates_sim}
        \centering
        \begin{tabular}{l l c c c c c c c}
            \toprule
            \multirow{2}{*}{\textbf{Demonstrations}} & \multirow{2}{*}{\textbf{Method}} & \multicolumn{6}{c}{\textbf{Task}}\\
            \cmidrule(lr){3-8}
            & & OpenDrawer & PushButtons & StackWine & SlideBlock & SweepToDustpan & PlaceCups\\
            \midrule
            100 & 3D Diffuser Actor~\cite{ke20243d} & 0.90 & 0.98 & 0.93 & \textbf{0.98} & 0.84 & 0.24 \\
            & Diffusion Policy~\cite{chi2023diffusionpolicy} & \textbf{0.96} & 0.00 & 0.42 & 0.31 & 0.32 & 0.00 \\
            & LSTM~\cite{hochreiter1997long} & 0.00  & 0.06 & 0.00 & 0.00 & 0.00 & 0.00 \\
            \cmidrule{1-8}
            5 & Diffusion Policy & 0.16 & 0.00 & 0.00 & 0.05 & 0.40 & 0.00 \\
            & LSTM & 0.00 & 0.00 & 0.00 & 0.00 & 0.00 & 0.00 \\
            & \textbf{TAPAS-GMM} (Ours) & 0.90 & \textbf{1.00} & \textbf{1.00} & 0.95 & \textbf{1.00} & \textbf{0.75} \\
            \bottomrule
        \end{tabular}
\end{table*}

\begin{table}
        \caption{Policy success without task-parameter selection or skill segmentation. TP-GMM uses neither. The other models ablate one factor each.}\label{tab:success_rates_abl}
        \centering
        \begin{tabular}{l c c c}
            \toprule
            & OpenDrawer &  SlideBlock & PlaceCups\\
            \midrule
            TAPAS-GMM (Ours) & \textbf{0.90} & \textbf{0.95} & \textbf{0.75} \\
            └─ w/o segmentation & 0.06 & 0.68 & 0.00  \\
            └─ w/o TP selection & 0.62 & 0.66 & 0.03 \\
            └─ Global TP selection & 0.62 & 0.66 & 0.10 \\
            \cmidrule{1-4}
            TP-GMM & 0.05 & 0.64 & 0.00 \\
            \bottomrule
        \end{tabular}
       \vspace{-0.2cm}
\end{table}

\begin{table}
        \caption{\rebuttal{Training and inference times across RLBench tasks.
        Inference times are normalized by the prediction horizon of the policy.
        }}\label{tab:inf_times}
        \centering
        \begin{tabular}{l c c c}
            \toprule
             & \multicolumn{2}{c}{\textbf{Training Duration}~[\unit{\hour}]} & \multirow{2}{*}{\textbf{Inference}~[\unit{\micro\second}]} \\
             \cmidrule(lr){2-3}
            &  5 Demos & 100 Demos &  \\
            \midrule
            TAPAS-GMM  & \textbf{0:00 {--} 0:01} & \textbf{00:02 {--} 00:11} & \textbf{\phantom{000}100} \\
            TP-GMM & 0:00 {--} 0:03 & 00:05 {--} 02:00 & \phantom{000}100  \\
            LSTM & 0:40 {--} 1:00 & 02:00 {--} 04:30 & \phantom{00}7.000\\
            Diffusion Policy & 1:30 {--} 5:00 & 30:00 {--} 99:00 & 100.000\\
            \bottomrule
        \end{tabular}
       \vspace{-0.3cm}
\end{table}

\subsection{Time-Based Initialization}\label{sec:init}
EM is not guaranteed to converge to the \emph{global} optimum, but only to a local optimum~\cite{zeestraten2018programming}.
Consequently, its initialization is of critical importance.
\(K\)-means clustering is popular but does not guarantee a global optimum either~\cite{zeestraten2018programming}.
\rebuttal{In contrast,} we efficiently \rebuttal{divide} the trajectories along the time dimension \rebuttal{into \(K\) bins}.
\figref{fig:tempalign} illustrates that complex task demonstrations are usually not aligned temporally.
However, our skill segmentation proposed in \secref{sec:trajseg} aligns them \rebuttal{and samples the segments to a common length so that \(K\)-bins yields} a good initialization for EM.
The combination of skill segmentation and time-based initialization enables the EM procedure to arrive at good local optima while also achieving significantly faster convergence.
Next, we discuss how we leverage it to efficiently select task parameters, too.

\subsection{Task-Parameterization}\label{sec:framesel}
Previous work has usually assumed groundtruth access to relevant task-parameters~\cite{li2023task,calinon2016tutorial} or used infrared motion tracking devices~\cite{zeestraten2018programming}, both of which restrict the applicability of the method.
Therefore, we propose an approach to automatically select the relevant task parameters from RGB-D observations.
Moreover, different skills in the same task require different information,
e.g., in \emph{pick and place}, the picking skill is often independent of the placing site.
Modeling the full task using both task parameters is not only more computationally expensive but also leads to worse task performance.
Hence, our parameter selection method has a synergistic effect with the skill segmentation proposed in \secref{sec:trajseg}.
We offer a modular two-staged approach to candidate generation based on visual features and task-parameter selection based on statistical considerations. 
Both operate independently of each other, rendering them straightforward to substitute.

\subsubsection{Candidate Generation}\label{sec:cand_gen}
Various methods exist for extracting task parameters from visual observations~\cite{chisari2023centergrasp}.
Due to their generalization capabilities, we build on recent advances in 3D scene keypoints~\cite{vonhartz2023treachery} and leverage a DINO vision transformer~\cite{amir2021deep}.
We sample a set of reference descriptors from the embedding of the first image observation of task demonstrations, either manually in a GUI or automatically using SAM-generated object masks~\cite{kirillov2023segment}.
The associated keypoints constitute the set of candidate task parameters.
We then localize them in each demonstration by establishing correspondences via the embedding distances~\cite{vonhartz2023treachery, amir2021deep}.
\rebuttal{In our experiments, we then set the \emph{orientation} of all keypoint frames to the identity quaternion, which we found sufficient for solving all evaluated tasks.
If the precise orientation of the keypoint frames should be required, it can be estimated using off-the-shelf methods~\cite{chisari2023centergrasp}}.

\subsubsection{Candidate Selection}\label{sec:cand_selec}
We select task parameters \emph{per skill} based on how well they explain the set of demonstrations.
For each skill and candidate frame, we first fit a single-frame model. %
We only perform the time-based initialization described in \secref{sec:init} and not the full EM to ensure temporal alignment across candidates and to speed up the procedure.
We then estimate the relative relevance of the per-frame models using the determinants of their Gaussian precision matrices, a metric astutely proposed by Alizadeh~\etal{}~\cite{alizadeh2014learning}.
Due to the temporal alignment, the relative precision over Gaussian components yields the relative relevance of the candidate frames over time.
By calculating the maximum over all Gaussian components, we select all frames that are relevant \emph{at some point} in the segment.
Let \({\scriptstyle \left(\boldsymbol\Sigma_k^f\right)^{-1}}\) denote the \(k\)-th component's precision matrix in the model of frame \(f\).
For \(C\) candidate frames with \(K\) Gaussians, we estimate
\begin{equation}\label{eq:frame_sel}
    \omega(f) = \max_{k=1}^K \frac{\det \left(\boldsymbol\Sigma_k^f\right)^{-1}}{\sum_{c=1}^C \det \left(\boldsymbol\Sigma_k^c\right)^{-1}}
\end{equation}
and select all frames \(f\) over some threshold \(\omega(f)>\tau\).

In contrast, Alizadeh~\etal{} fit a joint model to all candidates and the full task trajectory and calculate the relative precision during trajectory reconstruction~\cite{alizadeh2014learning}.
Our per-skill single-frame approach scales to long-horizon tasks and large sets of candidate parameters, while the temporal alignment eliminates the need for a reconstruction phase.
\begin{table*}
        \caption{Time-driven vs.\ state-driven \textbf{TAPAS-GMM}, with and without disturbance of the rollout.
        Among the time-driven policies we compare different post-processing strategies
        Among the state-driven policies, we compare naive and factorized velocities.
        Trajectory length is mean and standard deviation.}\label{tab:success_rates_txdx}
        \centering
        \begin{tabular}{l l c c c c c c c c}
            \toprule
             \multirow{3}{*}{\textbf{Driver}} & \multirow{3}{*}{\textbf{Variant}} & \multicolumn{4}{c}{\textbf{Policy success rate} $\uparrow$} & \multicolumn{4}{c}{\textbf{Successful trajectory length}~[steps] $\downarrow$} \\
            \cmidrule(lr){3-6}
            \cmidrule(lr){7-10}
            & & \multicolumn{2}{c}{Undisturbed} & \multicolumn{2}{c}{Disturbed} & \multicolumn{2}{c}{Undisturbed} & \multicolumn{2}{c}{Disturbed} \\
            \cmidrule(lr){3-4}
            \cmidrule(lr){5-6}
            \cmidrule(lr){7-8}
            \cmidrule(lr){9-10}
            & & PushBlock &  LiftBottle & PushBlock &  LiftBottle & PushBlock &  LiftBottle & PushBlock &  LiftBottle\\
            \midrule
            Time & None & 0.85 & 0.18 & 0.00 & 0.00 & $127 \pm 01$ & $239 \pm 09$ & - & - \\
             & Threshold & \textbf{0.96} & \textbf{0.79} & 0.00 & 0.00 & $149 \pm 01 $ & $310 \pm 06$ & - & - \\
             & TOPP-RA & 0.35 & 0.78 & 0.00 & 0.17 & $370 \pm 64$ & $426 \pm 26$ & - & $444 \pm 18$ \\
            \cmidrule{1-10}
            State & Naive & 0.76 & 0.18 & 0.76 & 0.19 & $206 \pm 71$ & $203 \pm 36$ & $266 \pm 59$ &  $454 \pm 37$ \\
             & Factorized & 0.83 & 0.33 & \textbf{0.83} & \textbf{0.32} & $141 \pm 53$ & $223 \pm 48$ & $228 \pm 47$ & $471 \pm 48$ \\
            \bottomrule
        \end{tabular}
\end{table*}

\begin{table*}
        \caption{Real world policy success rates.}\label{tab:success_rates_real}
        \centering
        \begin{tabular}{l c c c c c c c }
            \toprule
            & PickAndPlace &  PourDrink & MakeCoffee & SweepBlocks & OpenCabinet & StoreInDrawer & BakeBread \\
            \midrule
            LSTM & 0.00 & 0.00 & 0.00 & 0.00 & 0.00 & 0.00 & 0.00 \\
            Diffusion Policy & 0.00 & 0.00 & 0.00 & 0.00 & 0.00 & 0.00 & 0.00 \\
            TP-GMM & 0.20 & 0.08 & 0.00 & 0.16 & 0.00 & 0.68 & 0.36 \\
            TAPAS-GMM (Ours) & \textbf{1.00} & \textbf{1.00} & \textbf{0.96} & \textbf{0.96} & \textbf{0.92} & \textbf{1.00} & \textbf{0.96} \\
            \bottomrule
        \end{tabular}
       \vspace{-0.3cm}
\end{table*}

\section{Experimental Results}
We first evaluate the efficacy of \textit{TAPAS-GMM} for policy learning on a set of challenging manipulation tasks from the well established RLBench benchmark~\cite{james2019rlbench} and ablate the major components of our method.
Afterward, we compare time-driven and state-driven policies on custom tasks implemented in ManiSkill2~\cite{gu2022maniskill2}.
Finally, we verify our results on a real robot and demonstrate our method's generalization capabilities.
\figref{fig:tasks} illustrates the evaluated RLBench tasks and real-world tasks.
Additional details for all the experiments are provided in \secref{sec:exp_details} of the supplementary material.

\subsection{Time-Driven Policy Learning in Simulation}
We compare against LSTM~\cite{hochreiter1997long} and Diffusion Policy~\cite{chi2023diffusionpolicy} trained with 100 and 5 demonstrations.
In contrast, \textit{TAPAS-GMM} is only trained on five demonstrations.
To experimentally disentangle policy learning and representation learning, we use groundtruth object poses as inputs for the baselines and as \emph{candidate} task parameters for \textit{TAPAS-GMM}.
We evaluate our proposed \emph{visual} features in \secref{sec:real_world_exp} and \secref{sec:real_world_gen}.
We further compare against 3D Diffuser Actor~\cite{ke20243d}, which at the time of writing, is the incumbent on RLBench.
Note that it leverages a 3D scene representation, not object poses.
We evaluate all policies for 200 episodes.
\tabref{tab:success_rates_sim} shows that \textit{TAPAS-GMM} matches or outperforms the SOTA across all tasks while requiring significantly fewer demonstrations.
Moreover, in contrast to all baselines, \textit{TAPAS-GMM} scales to more complex long-horizon tasks such as \texttt{SweepToDustpan} and \texttt{PlaceCups}, without loss in performance.
Moreover, unlike the Diffusion Policy and LSTM, it excels when high precision is required, e.g., in \texttt{PushButton}.
\rebuttal{\tabref{tab:inf_times} reports training and inference times.
All GMMs are fitted on a CPU.
The deep learning baselines are trained on a NVIDIA RTX A6000.}

In \tabref{tab:success_rates_abl}, we ablate the relevance of task-parameter selection and skill segmentation.
The \emph{global} TP selection chooses the parameters for the full task jointly, instead of per skill.
As hypothesized, the task-parameter selection is more crucial on multi-object tasks with clutter (\texttt{PlaceCups}) than on single-object tasks (\texttt{OpenDrawer}, \texttt{SlideBlock}).
Across all tasks, we observe that the initial end-effector frame is irrelevant after aligning the gripper with the object and reduces policy success if still taken into account.
Skill segmentation is most crucial if the task is complex (\texttt{PlaceCups}) or precise grasps are required (\texttt{OpenDrawer}).
The large performance gap between \textit{TAPAS-GMM} and all other models on \texttt{PlaceCups} highlights the synergistic effect of skill segmentation and parameter selection.

\begin{figure}[tb]
    \centering
    \includegraphics[width=0.14\textwidth,valign=t]{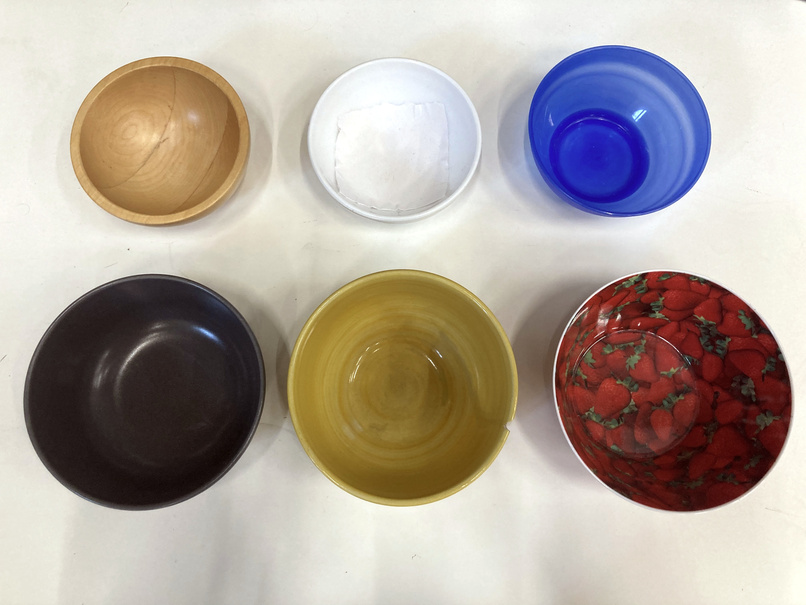}\hspace{0.1cm}%
    \includegraphics[width=0.105\textwidth,valign=t]{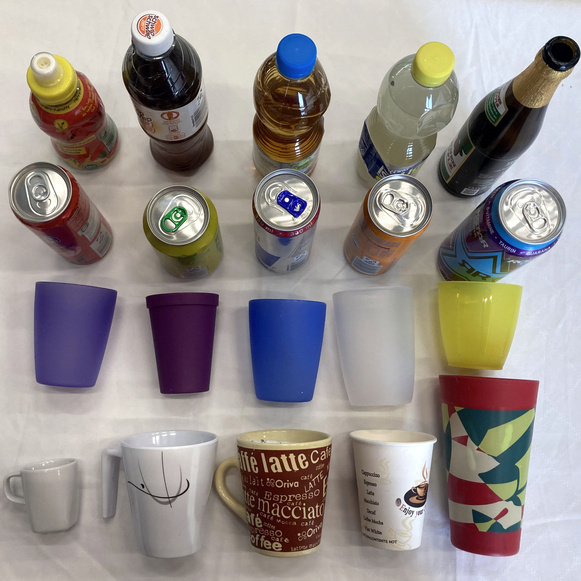}\hspace{0.1cm}%
    \includegraphics[width=0.105\textwidth,valign=t]{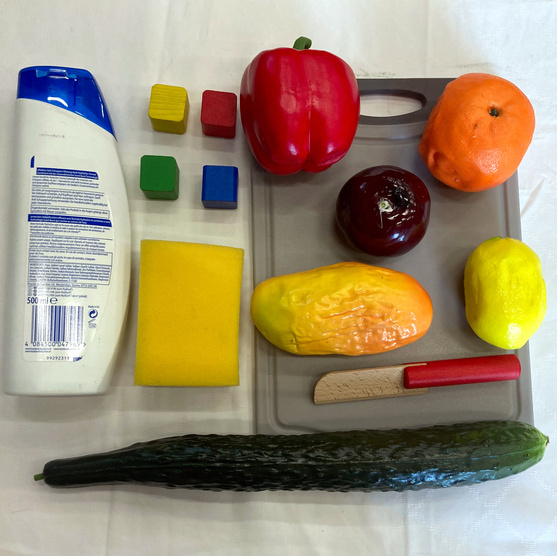}\hspace{0.1cm}%
    \includegraphics[width=0.105\textwidth,valign=t]{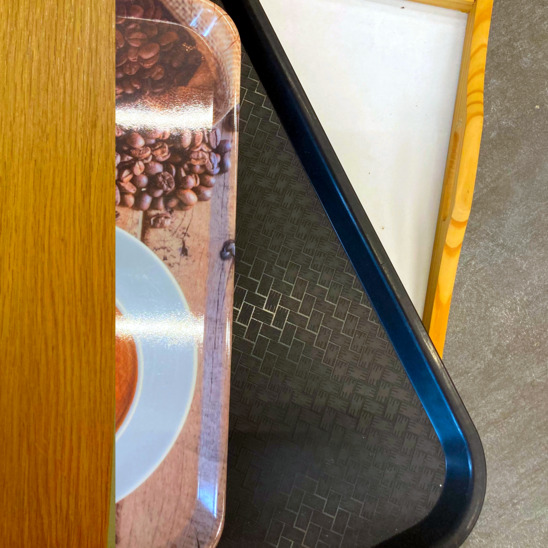}\\
    \vspace{0.1cm}
    {\footnotesize 
    \begin{tabular}{l c c c}
        \toprule
        & \multicolumn{3}{c}{\textbf{Variation}} \\
        \cmidrule(lr){2-4}
        \textbf{Task}        & {Object Instance} & {Clutter Objects} & {Environment} \\
        \midrule
        PickAndPlace    & 5/5        & 10/10         & 5/5             \\
        PourDrink       & 17/18      & 10/10        & 5/5             \\
        \bottomrule
    \end{tabular}
    }
    \caption{
    \rebuttal{Objects and success rates for our generalization study.
    \textit{Left:} object instances for both tasks, \textit{Center Right:} clutter objects, \textit{Right:} environments.
    }
}\label{fig:obj_gen}
   \vspace{-0.3cm}
\end{figure}

\subsection{Time-Driven vs.\ State-Driven Policies}\label{sec:state_res}
We compare the effectiveness of time-driven and state-driven GMMs in \tabref{tab:success_rates_txdx}.
To simulate disturbances during policy rollout, we reset the end-effector pose back to the starting pose, around the object contact point in the trajectory.
While state-based policies are harder to learn from few demonstrations, they outperform time-based policies when confronted with disturbances due to their time-invariance.
Furthermore, our action-factorized policies consistently outperform the naive implementation.

Moreover, time-based GMMs do not take into account the velocity limits of the robot.
Consequently, the predicted trajectories might not be feasible and require some post-processing, such as applying TOPP-RA~\cite{pham2018new} or only stepping the time when the current pose is close enough to the last prediction (thresholding).
These post-processing methods have their own strengths and weaknesses.
For example,\ TOPP-RA helps with precise grasping in \texttt{LiftBottle} and improves robustness against disturbances, but struggles with the drag of the object in \texttt{PushBlock}.
In contrast, our state-driven GMMs explicitly model the robot's velocity distribution in the demonstrations, thus not requiring post-processing.
Often, they also produce shorter trajectories.
The biggest drawback of state-driven GMMs that we observed experimentally, is their propensity for divergence during inference.
Under velocity control, pose errors can accumulate and push the policy outside the observed state space.
From this they often fail to recover, as frequently happens in \texttt{LiftBottle}.

\subsection{Real-World Policy Learning}\label{sec:real_world_exp}
We verify \textit{TAPAS-GMM} capabilities extensively on a real Franka Emika robot.
To this end, we construct a range of challenges such as long-horizon tasks up to 1000 time steps (e.g.\ \texttt{MakeCoffee}, \texttt{BakeBread}), tasks requiring high precision (e.g.\ \texttt{MakeCoffee}, \texttt{BakeBread}), and articulated objects (e.g.\ \texttt{OpenCabinet}, \texttt{StoreInDrawer}).
All policies are trained on five task demonstrations and evaluated for 25 episodes.
We generate candidate task parameters from wrist-camera observations as proposed in \secref{sec:cand_gen}.

The policy success rates in \tabref{tab:success_rates_real} are consistent with the results in our simulated experiments and demonstrate the core advantages of our approach:
\begin{enumerate*}
    \item strong generalization from few demonstrations,
    \item handling of articulated objects, and
    \item scaling to complex tasks and long horizons.
\end{enumerate*}
Consistent with our simulated ablation experiment, TP-GMMs struggle with pose accuracy and gripper timing due to locally irrelevant task parameters and poor temporal alignment. Moreover, the deep learning baselines fail to generalize from five demonstrations and do not learn to solve any task instance. In contrast, \textit{TAPAS-GMM} effortlessly scales to long-horizon tasks, without any loss in accuracy or generalization performance.

\subsection{Visual Generalization and Skill Reuse}\label{sec:real_world_gen}
Prior work has extensively demonstrated the generalization capabilities of keypoints as visual features~\cite{ vonhartz2023treachery}, which our policies inherit through the parameter generation proposed in \secref{sec:cand_gen}.
We study the resulting zero-shot generalization capabilities of our policies on \texttt{PickAndPlace} and \texttt{PourDrink} by evaluating the policy \rebuttal{on a set of} unseen objects instances, clutter objects, and task environments.
\rebuttal{\emph{TAPAS-GMM}} exhibits strong generalization performance across the board\rebuttal{, as shown in \figref{fig:obj_gen}}.
We further provide qualitative results in the supplementary video.

Another advantage of \textit{TAPAS-GMM} is that it learns multiple skill models, which can be recombined to solve novel tasks.
In the supplementary video, we demonstrate how the segmented skills can be sequenced and reused in novel ways.
In particular, the \texttt{BakeBread} task consists of taking the tray out of the oven and placing the bread on it.
As an example, we invert and reorder these skills to construct a policy that takes the bread from the tray and inserts the tray into the oven.

\section{Conclusion}
We present \textit{TAPAS-GMM}, a method for learning complex manipulation tasks from only five demonstrations.
By segmenting skills and automatically selecting relevant task parameters per skill, our method scales both tasks with long horizons and many parameters.
Through leveraging visual semantic features, our method further generalizes across object instances, environments, and clutter.
Finally, our proposed velocity factorization makes time-invariant task-parameterized policies more effective.

{\parskip=3pt
\noindent\textit{Limitations}: 
\textit{TAPAS-GMM} does not reason about kinematic limits or object collisions.
Also, achieving decent task space coverage from only five demonstrations can be challenging with multimodal trajectory distributions. 
This is evident in the \texttt{PlaceCups} task, where workspace limits lead to vastly different trajectory shapes across task instances.}

\typeout{}
{\footnotesize
\bibliographystyle{IEEEtran}
\bibliography{root}}

\clearpage
\renewcommand{\baselinestretch}{1}
\setlength{\belowcaptionskip}{0pt}

\begin{strip}
\begin{center}
\vspace{-5ex}
\textbf{\LARGE \bf
The Art of Imitation: Learning Long-Horizon Manipulation Tasks from Few Demonstrations\\}
\vspace{3ex}
\Large{\bf- Supplementary Material -}\\
\vspace{0.4cm}
\normalsize{Jan Ole von Hartz, Tim Welschehold, Abhinav Valada, and Joschka Boedecker}
\end{center}
\end{strip}

\setcounter{section}{0}
\setcounter{equation}{0}
\setcounter{figure}{0}
\setcounter{table}{0}
\setcounter{page}{1}
\makeatletter

\renewcommand{\thesection}{S.\arabic{section}}
\renewcommand{\thesubsection}{S.\arabic{subsection}}
\renewcommand{\thetable}{S.\arabic{table}}
\renewcommand{\thefigure}{S.\arabic{figure}}
\renewcommand{\theequation}{S.\arabic{equation}}

\section{Additional GMM Details}\label{sec:gmreqs}
\subsection{Full GMR Equations}
\rebuttal{For inference, we employ Gaussian Mixture Regression (GMR)~\cite{cohn1996active} to estimate $p_\mathcal{G} (\boldsymbol o\mid\boldsymbol i)$.
In a GMM without transition model this is given by}
\begin{equation}\label{eq:gmr_gmm}
    p_\mathcal{G} (\boldsymbol o\mid\boldsymbol i) = \sum_{k=1}^K \pi_k^{\boldsymbol i} \cdot\mathcal{N}(\boldsymbol o\mid \boldsymbol\mu_k ^{o\mid i}, \boldsymbol\Sigma_k^{o\mid i}).
\end{equation}
Let
\begin{equation*}
    \boldsymbol\mu_k = \begin{psmallmatrix}\boldsymbol\mu^\mathcal{I}_k\\\boldsymbol\mu^\mathcal{O}_k\end{psmallmatrix} \text{ and } \boldsymbol\Sigma_k = \begin{psmallmatrix}\boldsymbol\Sigma^{\mathcal{I}\mathcal{I}}_k & \boldsymbol\Sigma^{\mathcal{I}\mathcal{O}}_k\\\boldsymbol\Sigma^{\mathcal{O}\mathcal{I}}_k & \boldsymbol\Sigma^{\mathcal{O}\mathcal{O}}_k\end{psmallmatrix}
\end{equation*}
be the partition of \(\boldsymbol\mu_k\) and \(\boldsymbol\Sigma_k\) into the input and output dimensions.
Then, for \(\boldsymbol i\in\mathcal{I}\) and \(\boldsymbol o\in\mathcal{O}\), we have:
\begin{equation}\label{eq:gmrsupp}
    p_\mathcal{G} (\boldsymbol o\mid\boldsymbol i) = \sum_{k=1}^K \pi_k^{i} \cdot\mathcal{N}(\boldsymbol o\mid \boldsymbol\mu_k ^{o\mid i}, \boldsymbol\Sigma_k^{o\mid i}),
\end{equation}
where
\begin{equation}
\begin{split}
    \pi_k^{\boldsymbol i} &= \frac{\mathcal{N}(\boldsymbol i\mid \boldsymbol\mu_k^\mathcal{I}, \boldsymbol\Sigma_k^\mathcal{I})}{\sum_{l=1}^K \mathcal{N}(\boldsymbol i\mid \boldsymbol\mu_l^\mathcal{I}, \boldsymbol\Sigma_l^\mathcal{I})},\\
    \boldsymbol\mu_k^{o\mid i} &= \boldsymbol\mu_k^o + \boldsymbol\Sigma_k^{oi}(\boldsymbol\Sigma_k^{ii})^{-1}(\boldsymbol i - \boldsymbol\mu_k^i),\\
    \boldsymbol\Sigma_k^{o\mid i} &= \boldsymbol\Sigma_k^{oo} - \boldsymbol\Sigma_k^{oi}(\boldsymbol\Sigma_k^{ii})^{-1}\boldsymbol\Sigma_k^{io} .   
\end{split}
\end{equation}

\rebuttal{\subsection{Hidden Markov Model}
The GMM's posterior \(p_\mathcal{G}(\boldsymbol o\mid \boldsymbol i)\) in \eqref{eq:gmr_gmm} only depends on the priors \(\pi_k^{i}\) and the conditioned Gaussians, not on any previous observations.
The Hidden Markov Model (HMM)~\cite{calinon2010hmm} augments the GMM by associating each Gaussian with a state \(k\) and modeling the transition probability \(a_{kl}\) from state \(k\) to state \(l\).
I.e.\ \(\mathcal{H}=\{\pi_k, \{a_{kl}\}_{l=1}^K, \boldsymbol\mu_k, \boldsymbol\Sigma_k\}_{k=1}^K\).
This helps the model to differentiate between components that are close together in the input space, yet model different parts of a task or movement.
For a sequence of observations \(\{\boldsymbol i_t\}_{t=0}^T\), the computation of the prior thus changes to 
\begin{equation}\label{eq:hmm}
    \pi_k^{i_t} = \frac{\left(\sum_{l=1}^K \pi_l^{i_{t-1}} a_{lk}\right)\mathcal{N}(\boldsymbol i_t\mid \boldsymbol\mu_k^\mathcal{I}, \boldsymbol\Sigma_k^{\mathcal{I}})}{\sum_{m=1}^K\left[\left(\sum_{l=1}^K \pi_l^{i_{t-1}} a_{lm}\right)\mathcal{N}(\boldsymbol i_t\mid \boldsymbol\mu_m^\mathcal{I}, \boldsymbol\Sigma_m^{\mathcal{I}}) \right]}.
\end{equation}}

\rebuttal{For clarity, we continue referring to our models as GMMs, even though \emph{all} of them use the HMM extension.}

\subsection{Task Parameterization}
\rebuttal{For task-parameterization, we need to estimate the per-frame marginal model \(p_f(\mathcal{I}, \mathcal{O} \mid\mathcal{G})\) for each frame \(1\le f\le F\), which we define as
\begin{equation}\label{eq:marg}
    p_f(\mathcal{I}, \mathcal{O} \mid\mathcal{G}) = \left\{\pi_k^f, \boldsymbol\mu_k^f, \boldsymbol\Sigma_k^f\right\}_{k=1}^K.
\end{equation}
For the time-driven model, it is given by
\begin{equation}
    p_f(\mathcal{I}, \mathcal{O} \mid\mathcal{G}) = p(t, \boldsymbol x (f) \mid \mathcal{G}).
\end{equation}
In contrast, in the state-driven case we estimate
\begin{equation}
    p_f(\mathcal{I}, \mathcal{O} \mid\mathcal{G}) = p(\boldsymbol x (f), \dot{\boldsymbol x} (f) \mid \mathcal{G}).
\end{equation}}

\rebuttal{The instantiation of a task parameter \(f\) is given by its origin \(\boldsymbol b_f\) and rotation matrix \(\boldsymbol A_f\).
To adapt to a given set of task parameters \(\Theta=\{\boldsymbol A_f, \boldsymbol b_f\}_{f=1}^F\), we transform the per-frame marginals to the world frame as per \eqref{eq:frametrans}.
Note, that in the frame transformation we have to account for the extra dimensions of the GMM as well.
These include the time dimension (in the time-driven case) and global action dimensions, such as gripper actions.
For these we apply an identity transform with \(\boldsymbol{\tilde{A}}_f = \boldsymbol I\), \(\boldsymbol{\tilde{b}} = \boldsymbol 0\).
For the velocity-dimensions, we assume static frames, hence transforming with \(\boldsymbol{\tilde{A}}_f = \boldsymbol A_f\), \(\boldsymbol{\tilde{b}} = \boldsymbol 0\).}

The order of marginalization, \eqref{eq:marg}, transformation, \eqref{eq:frametrans}, product, \eqref{eq:join}, and GMR, \eqref{eq:gmr}, can be altered.
Some authors perform GMR \emph{per} marginal and then join the predicted Gaussians~\cite{zeestraten2018programming}.
However, the order proposed here facilitates the sequencing of HMMs, that we introduce in \secref{sec:sequence}.

\section{Additional Manifold Details}\label{sec:mandetails}
\subsection{\texorpdfstring{$\mans^3$}~Logarithmic Map}
Zeestraten~\cite{zeestraten2018programming} proposes to use the following logarithmic map for $\mans^3$
\begin{equation}
    \text{Log}_{\mathbf{e}}(\mathbf{g}) = \begin{cases}
        \arccos^*(q_0)\frac{\textbf{q}}{\lVert\textbf{q}\rVert} &, q_0\neq 1\\
        \begin{bmatrix}0, 0, 0\end{bmatrix}^T &, q_0=1.
    \end{cases}
\end{equation}
Here, \(q_0\) denotes the quaternion's real part.
They further define
\begin{equation}
    \arccos^*(\rho) = \begin{cases}
        \arccos(\rho) - \pi &, -1\le\rho< 0\\
        \arccos(\rho) &, 0\le\rho\le 1.
    \end{cases}
\end{equation}
This part-wise definition addresses the double-covering of rotations by \(\mans^3\) so that antipodal rotations on \(\mans^3\) have zero distance in tangent space.
However, it introduces discontinuities at \(\rho = 0\), which we found problematic when trying to find local near-linear segments.
Instead, we use the standard \(\arccos\) function and preprocess our quaternion data to ensure continuity such that no two successive rotations are antipodal.

\begin{figure}[tb]
  \centering
  \includegraphics[width=\columnwidth]{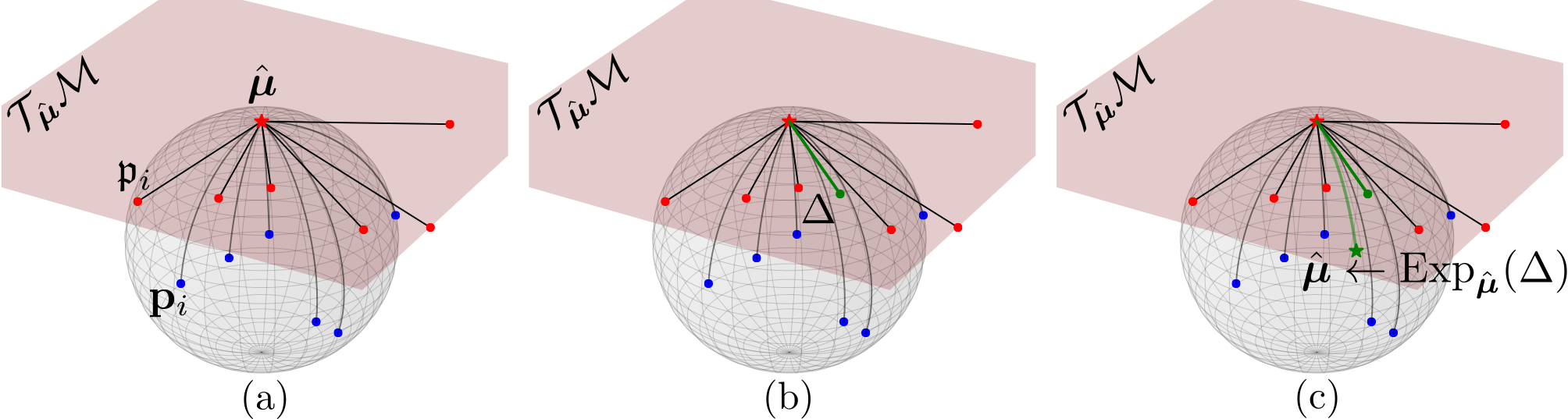}
  \caption{Iterative likelihood maximization for finding a Riemannian center of mass.
  (a) Given a current estimate \(\hat{\boldsymbol\mu}\) ($\textcolor{red}{\filledstar}$), project the data points \(\boldsymbol p_i\) ($\textcolor{blue}{\bullet}$) in the tangent space ($\textcolor{red}{\bullet}$) as \(\mathfrak{p}_i=\text{Log}_{\hat{\boldsymbol\mu}}(\boldsymbol p_i)\).
  (b) Compute an update \(\Delta\) in tangent space ($\textcolor{svgGreen}{\bullet}$) and
  (c) project it back onto the manifold ($\textcolor{svgGreen}{\filledstar}$) using \(\text{Exp}_{\hat{\boldsymbol\mu}}\).
  Iterate until convergence.
  Source: Adapted from \cite{zeestraten2018programming}.
  }\label{fig:mean_iter}
 \vspace{-0.3cm}
\end{figure}

\rebuttal{
\subsection{Riemannian Manifolds}\label{sec:rieman_app}
Each Riemannian manifold \(\mani\) is differentiable, i.e., at each point \(\mathbf{x}\), it has a tangent space \(\tans_\mathbf{x}\mani\), that is a real vector space.
The \emph{logarithmic map} \(\text{Log}_{\mathbf{x}}: \mathcal{M}\to \mathcal{T}_{\mathbf{x}}\mathcal{M}\) straightens the manifold's curvature while preserving geodesic distance.
The inverse is the \emph{exponential map} \(\text{Exp}_{\mathbf{x}}: \mathcal{T}_{\mathbf{x}} \mathcal{M}\to \mathcal{M}\).
Both depend on the \emph{base} \(\mathbf{x}\), at which they are applied.
Following Zeestraten~\cite{zeestraten2018programming}, we formulate all previously Euclidean operations in EM and GMR as non-linear Maximum Likelihood Estimation problems.
For example, to compute the center of mass \(\boldsymbol\mu\) of a Riemannian Gaussian, we initialize the procedure with some estimate \(\hat{\boldsymbol\mu}\) and solve the MLE using the following iterative process.
\begin{enumerate}
    \item Tangent project the data using \(\text{Log}_{\hat{\boldsymbol\mu}}\).
    \item Update \(\Delta\in\tans_{\hat{\boldsymbol\mu}} \mani\) using a Gauss-Newton step. 
    \item Back-project the update \(\hat{\boldsymbol\mu}\leftarrow\text{Exp}_{\hat{\boldsymbol\mu}}(\Delta)\).
\end{enumerate}
In practice, only a handful of iterations suffice for \(\hat{\boldsymbol\mu}\) to converge.
\figref{fig:mean_iter} illustrates the procedure.
All other operations such as conditioning can be solved in a similar manner~\cite{zeestraten2018programming}.
Finally, we use the concept of parallel transport to implement the affine frame transformations in \eqref{eq:frametrans} for the full pose.
We provide details on this procedure next. 
}

\subsection{\texorpdfstring{$\mans^n$ Frame Transformation}{Sn frame transformations}}
Recall from \secref{sec:tpgmm} that for a given instance of task parameters \(\Theta=\{\boldsymbol A_f, \boldsymbol b_f\}_{f=1}^F\), we transform the per-frame marginals to the world frame using  \eqref{eq:frametrans}.
However, \eqref{eq:frametrans} only applies to Euclidean Gaussians.
In contrast, for Riemannian Gaussians~\cite{zeestraten2018programming}, we have
\begin{align}\label{eq:rietrans}
    \hat{\boldsymbol\mu}^f_k &= \text{Exp}_{\boldsymbol b_f}\left(\boldsymbol A_f\text{Log}_\mathbf{e}\left({\boldsymbol\mu}^f_k\right)\right) \\
    \hat{\boldsymbol\Sigma}^f_k &= \left(\boldsymbol A_f \boldsymbol\Sigma_k^f\boldsymbol A_f^T\right)_{\Vert_{{\boldsymbol b}_f}^{\hat{\boldsymbol\mu}^f_k}}.\label{eq:rietrans2}
\end{align}
The subscript ${\scriptstyle \Vert_{{\boldsymbol b}_f}^{\hat{\boldsymbol\mu}^f_k}}$ denotes the \emph{parallel transport} of the covariance matrix from \({\boldsymbol\mu}^f_k\) to \(\hat{\boldsymbol\mu}^f_k\) using \({\boldsymbol b_f}\)~\cite{zeestraten2018programming}.
We discuss the role of parallel transport in more detail in \secref{sec:covtrans}.
For \(\mans^3\) we follow prior work~\cite{zeestraten2018programming} and transform the quaternion manifold using \(\boldsymbol A_f = \boldsymbol I_3\) and \(\boldsymbol b_f = \boldsymbol q_f\), where \(\boldsymbol q_f\) is the rotation quaternion that is equivalent to the rotation matrix \(\boldsymbol A_f\).
Recall that we use the \(\mans^2\) manifold to model the direction of actions (velocities).
To enable applying the same frame transformation as for \(\mans^3\), we modify the tangent action function to accept quaternion arguments.
Note that any element \(\mathbf{p}\in\mans^2\) represents a three-dimensional vector of the unit norm, which we can represent as a vector quaternion \(\begin{bmatrix}0, \mathbf{p}\end{bmatrix}^T\in\mans^3\).
In this way, we can rotate a manifold element \(\mathbf{p}\in\mans^2\) by a quaternion \(\boldsymbol q\in\mans^3\) via
\begin{equation}\label{eq:vecrot}
    \begin{bmatrix}0,\hat{\boldsymbol p}\end{bmatrix}^T = \mathbf{q} \begin{bmatrix}0, \mathbf{p}\end{bmatrix}^T \mathbf{q}^{-1},
\end{equation}
i.e., we embed the vector in \(\mans^3\), compute the Hamilton product, and drop the leading zero to map the resulting vector quaternion back, s.t.\ \(\hat{\boldsymbol p}\in\mans^2\).
Parallel transport moves an element \(\boldsymbol p_{\boldsymbol g}\) of the manifold from one tangent base \(\boldsymbol g\) to another \(\boldsymbol h\).
Specifically, ${\scriptstyle \Vert_{{\boldsymbol b}_f}^{\hat{\boldsymbol\mu}^f_k}}$ moves \(\boldsymbol \mu_k^f\) from the base \(\boldsymbol e\), which is the identity element of the manifold, to \(\boldsymbol b_f\), yielding \(\hat{\boldsymbol\mu}^f_k\).
As \(\boldsymbol b_f\not\in\mans^2\), we first need to define a surrogate element to use as the new base.
We define
\begin{equation}\label{eq:s2bhat}
    \hat{\boldsymbol b}_f = \left(\boldsymbol q_f \begin{bmatrix}0, \boldsymbol e\end{bmatrix}^T \boldsymbol q_f^{-1}\right)_{[1:3]}    
\end{equation}
and substitute \(\hat{\boldsymbol b}_f\in \mans^2\) for \(\boldsymbol b_f\in \mans^3\) in \eqref{eq:rietrans2}.
Moreover, the exponential map in \eqref{eq:rietrans} depends on the base at which it is applied.
For homogeneous manifolds, such as \(\mans^n\), it can be defined as
\begin{equation}
    \text{Exp}_{\boldsymbol g} (\mathfrak{p}_{\boldsymbol g}) = \mathcal{A}_{\boldsymbol e}^{\boldsymbol g}(\text{Exp}_{\boldsymbol e}(\mathfrak{p}_{\boldsymbol g})),
\end{equation}
where \(\mathfrak{p}_{\boldsymbol g}\) denotes the tangent element associated with the manifold element \(\boldsymbol p_{\boldsymbol g}\).
\(\mathcal{A}_{\boldsymbol g}^{\boldsymbol h}(\boldsymbol p_{\boldsymbol g})\) denotes the action function~\cite{zeestraten2018programming}, which maps \(\boldsymbol p_{\boldsymbol g}\) to \(\boldsymbol p_{\boldsymbol h}\) along a geodesic, such that pairwise distances between both \(\boldsymbol p_{\boldsymbol g}\) and \(\boldsymbol g\), and \(\boldsymbol p_{\boldsymbol h}\) and \(\boldsymbol h\) are identical.
As \(\mathcal{A}_{\boldsymbol g}^{\boldsymbol h}(\boldsymbol p_{\boldsymbol g})\) requires \(\boldsymbol g\) to be a manifold element, we again leverage \eqref{eq:vecrot} to calculate
\begin{align}
    \text{Exp}_{\tilde{\boldsymbol b}_f}(\mathfrak{p}_{\boldsymbol g}) &= \mathcal{A}_{\boldsymbol e}^{\tilde{\boldsymbol b}_f}(\text{Exp}_{\boldsymbol e} (\mathfrak{p}_{\boldsymbol g})) \\
    \begin{bmatrix}0,\mathcal{A}_{\boldsymbol e}^{\tilde{\boldsymbol b}_f}(\boldsymbol p_{\boldsymbol g})\end{bmatrix}^T  &= \boldsymbol q_f \begin{bmatrix}0, \boldsymbol p_{\boldsymbol g}\end{bmatrix}^T \boldsymbol q_f^{-1}.\label{eq:s2btilde}
\end{align}
For reference, the quaternion action on \(\mans^3\) is similarly defined as
\begin{align}
    \mathcal{A}_{\boldsymbol g}^{\boldsymbol h}(\boldsymbol p_{\boldsymbol g}) &= \boldsymbol h \boldsymbol g^{-1} \boldsymbol p_{\boldsymbol g} \\
    \mathcal{A}_{\boldsymbol e}^{\boldsymbol b_f}(\boldsymbol p_{\boldsymbol g}) &= \boldsymbol q_f  \boldsymbol p_{\boldsymbol g}.
\end{align}

Putting the modified parallel transport and action together, for \(\mans^2\) we have
\begin{align}
    \hat{\boldsymbol\mu}^f_k &= \text{Exp}_{\hat{\boldsymbol b}_f}\left(\boldsymbol A_f\text{Log}_\mathbf{e}\left({\boldsymbol\mu}^f_k\right)\right) \\
    \tilde{\boldsymbol\Sigma}^f_k &= \left(\boldsymbol A_f \boldsymbol\Sigma_k^f\boldsymbol A_f^T\right)_{\Vert_{\hat{\boldsymbol b}_f}^{\hat{\boldsymbol\mu}^f_k}},
\end{align}
with \(\hat{\boldsymbol b}_f\) and \(\text{Exp}_{\tilde{\boldsymbol b}_f}\) as defined in \eqref{eq:s2bhat} and \eqref{eq:s2btilde}, respectively.

\section{Covariance Regularization}\label{sec:covtrans}
\begin{figure}
    \centering
    \begin{subfigure}[t]{0.48\linewidth}
      \includegraphics[width=\columnwidth]{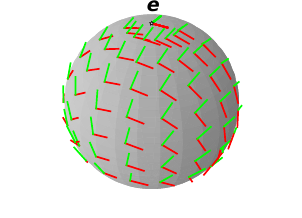}
      \caption{Tangent space alignment w.r.t.\ \(\boldsymbol e\).}\label{fig:tane}
   \end{subfigure}\hfill
    \begin{subfigure}[t]{0.48\linewidth}
      \includegraphics[width=\columnwidth]{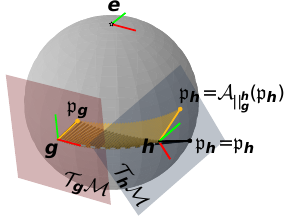}
        \caption{Base misalignment between \(\boldsymbol g\) and \(\boldsymbol h\).}\label{fig:transport}
   \end{subfigure}
  \caption{Parallel transport of the manifold element \(\mathfrak{p}\) from basis \(\boldsymbol g\) to basis \(\boldsymbol h\) requires a rotation by \(\mathcal{A}_{\Vert_g^h(\mathfrak{p})}\) to compensate for the misaligned tangent spaces.
  Source: Adapted from \cite{zeestraten2018programming}.
  }\label{fig:paralleltrans}
\end{figure}
\begin{figure*}
    \centering
    \begin{subfigure}[t]{\linewidth}
        \includegraphics[width=\linewidth]{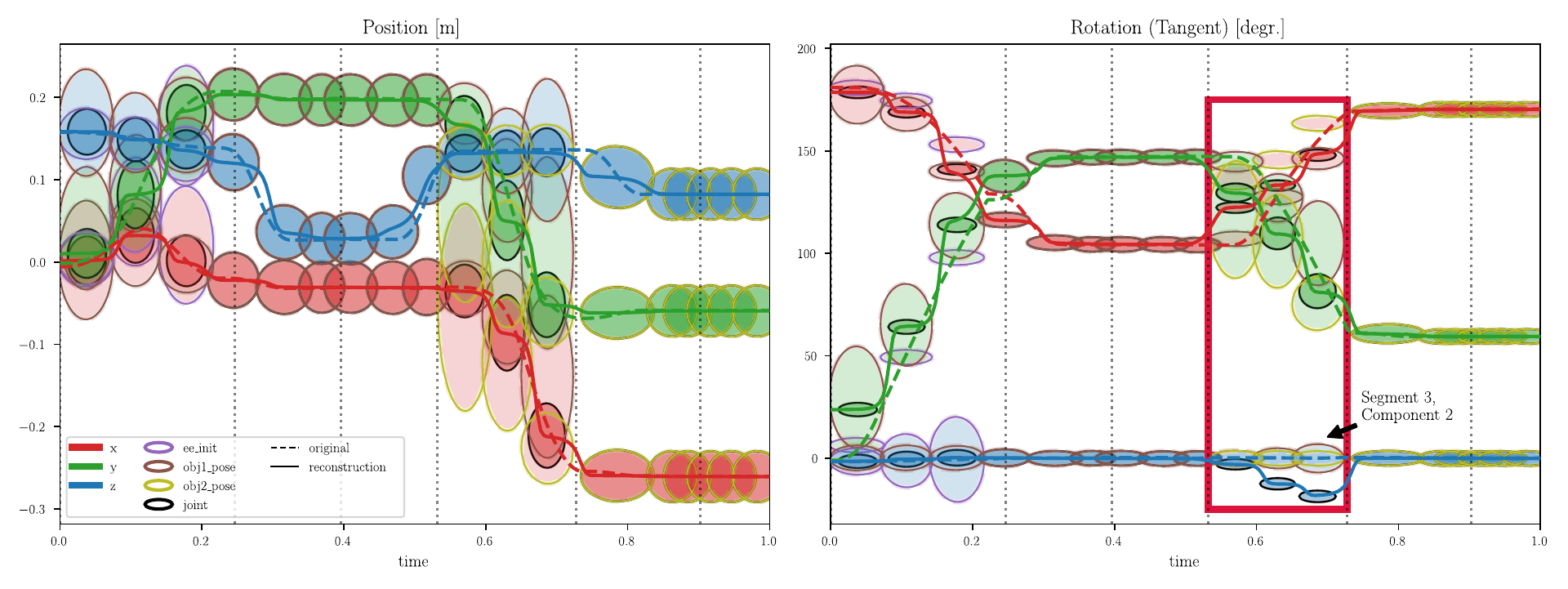}
        \caption{Faulty reconstruction due to unwanted correlations.
        The plot shows the model dimensions over time. Left: position as  \(x, y, z\) coordinates. Right: rotation in tangent space as scaled axis-angle components \(\mathit{qx}, \mathit{qy}, \mathit{qz}\).
        Vertical dotted lines indicate the borders of the segment models.
        Red, green, and blue represent the individual components.
        The ellipses with colored borders represent the transformed marginals.
        The ones with black borders represent the joint model, which is the product of the marginals.
        Note how the rotation reconstruction is faulty around \(t=0.6\).
        Note further, how at the same time the blue joint Gaussians are shifted down from where one would expect them.}\label{fig:faultrec}
    \end{subfigure}\\
    \begin{subfigure}[t]{0.52\linewidth}
        \includegraphics[width=\linewidth]{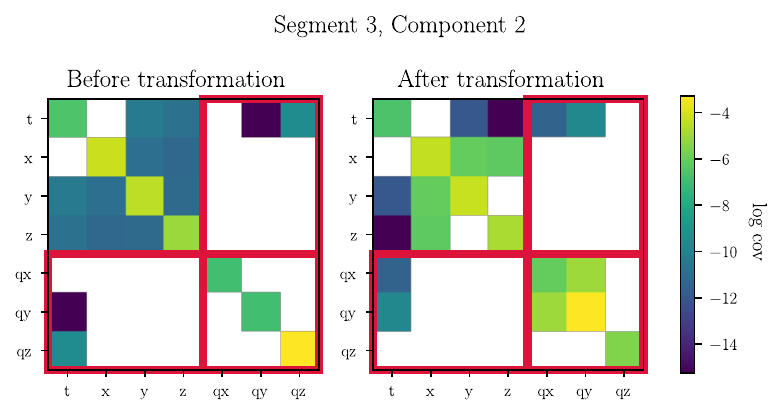}
        \caption{Log covariance of segment 3, component 2 of the marginal model of \emph{obj1} - before and after transformation. White values are near \(-\infty\). \(t\) indicates time, \(x, y, z\) position, \(\mathit{qx}, \mathit{qy}, \mathit{qz}\) rotation. Note how the transformation adds in additional correlations for the rotation dimensions.}\label{fig:faultcov}
    \end{subfigure}\hfil
    \begin{subfigure}[t]{0.22\linewidth}
        \includegraphics[width=\linewidth]{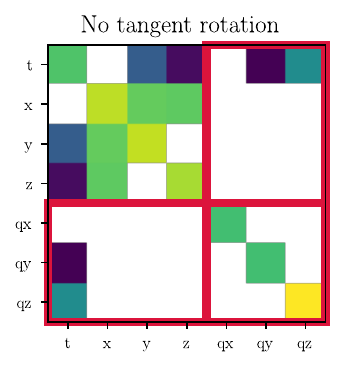}
        \caption{Without tangent rotation. Note that there are no novel correlations between the rotation dimensions.}\label{fig:norotcov}
    \end{subfigure}\hfil
    \begin{subfigure}[t]{0.22\linewidth}
        \includegraphics[width=\linewidth]{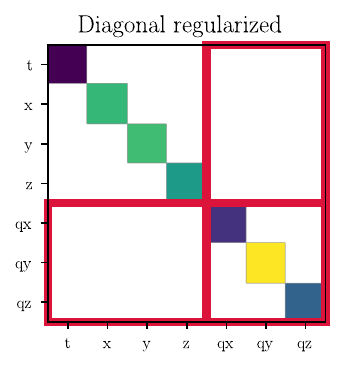}
        \caption{Diagonal regularization removes the novel unwanted correlations between the rotation dimensions.}\label{fig:fixedcov}
    \end{subfigure}\\
    \begin{subfigure}[t]{\linewidth}
        \includegraphics[width=\linewidth]{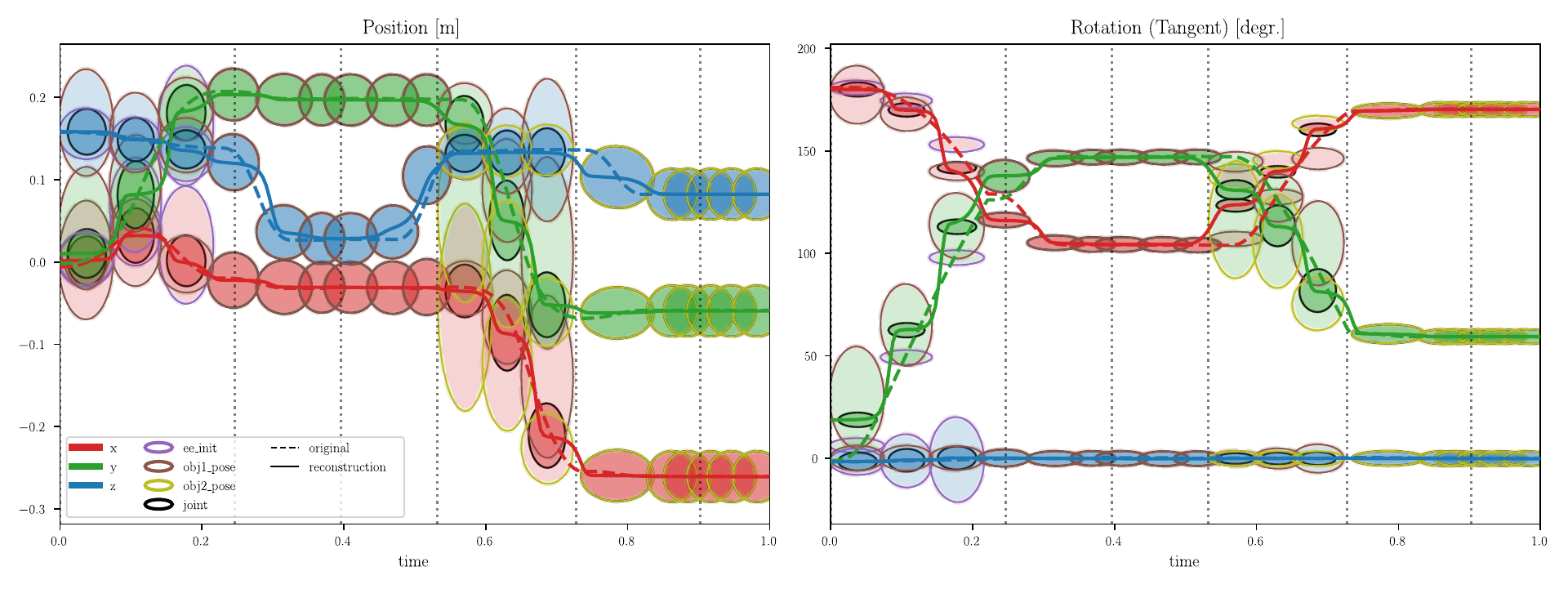}
        \caption{Fixed reconstruction after applying the additional diagonal regularization after the parallel transport of the covariance matrices.
        Note how the joint model's Gaussians for the rotation dimensions are now located at the expected positions.}\label{fig:fixedrec}
    \end{subfigure}
    \caption{Frame transformations introduce unwanted correlations, causing faulty reconstructions.
    \subfiguresubref{fig:faultrec} shows a faulty reconstruction for the rotation.
    This is caused by the correlations introduced by the frame transformation shown in \subfiguresubref{fig:faultcov}.
    \subfiguresubref{fig:norotcov} highlights that this is a direct consequence of the rotation \(\mathcal{A}_{\Vert_g^h(\mathfrak{p})}\) explained in \figref{fig:paralleltrans}.
    To remedy this issue, we apply an additional diagonal regularization after the parallel transport.
    \subfiguresubref{fig:fixedcov} shows the new covariance matrix and \subfiguresubref{fig:fixedrec} the fixed reconstruction.
    }\label{fig:covreg}
\end{figure*}

During EM, the model might learn unwanted correlations between some of the data dimensions.
This well know problem can lead to faulty predictions, as illustrated in \figref{fig:faultrec}.
Usually, this issue can be solved using shrinkage regularization in EM~\cite{zeestraten2018programming}.
However, we found the regularization of the EM to be sometimes insufficient to ensure high quality predictions in Riemannian TP-GMMs.
Specifically, the transformation of the local marginal models can re-add unwanted correlations.
To the best of our knowledge, this fact was not reported in prior work, so we discuss it in this section.

Consider a Riemannian TP-GMM on the manifold \(\mani=\mathbb{R}\times\mathbb{R}^3\times\mathcal{S}^3\), modeling time, as well as position and rotation of the end-effector.
Note that for the rotation quaternion, the covariance matrix is defined and computed in the tangent space \(\mathcal{T}_{\boldsymbol x}\mans^3\).
Thus, it is given by a \(3 \times 3\) matrix.
Recall the frame transformation of the marginal models discussed in \secref{sec:tpgmm}.
While regularization of EM can dampen or remove unwanted correlations, the frame transformation can add novel correlations back in, as shown in \figref{fig:faultcov}.
Perhaps surprisingly, the culprit is \emph{not} the transformation of the covariance matrix in \eqref{eq:frametrans}.
While it clearly changes the correlations between the (Euclidean) position dimensions \(x, y, z\), the novel correlation between the rotation dimensions \(qx\) and \(qy\) hails elsewhere.

Riemannian TP-GMMs use \emph{parallel transport} to implement frame transformations~\cite{zeestraten2018programming}.
Parallel transport moves a Gaussian from its current tangent basis (mean) to a new tangent basis along the connecting geodesic, as shown in \figref{fig:transport}.
Following the geodesic ensures compensation for the relative rotation between the old and new tangent space, indicated in \figref{fig:tane}.
In practice, this is implemented by first shifting the mean of the Gaussian to its new position, and then applying the compensating rotation matrix to the covariance matrix~\cite{zeestraten2018programming}.
This rotation of the covariance matrix introduces novel correlations.
For illustration, we have heavily regularized the marginal covariance matrix in \figref{fig:faultcov} before transforming it.
It does not contain any correlation between the rotation dimensions.
Note how the frame transformation adds in additional correlations between the rotation dimensions \(\mathit{qx}\) and \(\mathit{qy}\), leading to the faulty reconstruction in \figref{fig:faultrec}.
\figref{fig:norotcov} shows the same transformed covariance, but omits the compensating rotation of the tangent space.
Crucially, it does not contain any correlation between \(qx\) and \(qy\).

The reconstruction of \(\mathit{qz}\) is also affected because of two reasons.
First, recall the product of marginal models in \eqref{eq:join}.
If another marginal correlates, for example, \(\mathit{qy}\) and \(\mathit{qz}\) due to a different frame transformation, then in the joint model, all three rotation dimensions are now correlated.
This might be, why previous work computed the GMR per marginal and only joint the predicted Gaussians~\cite{zeestraten2018programming}.
However, combining the marginals first, homogenizes the input dimensions of GMR, thus simplifying the cascading of models, discussed in \secref{sec:sequence}.
Second, the rotation dimensions interact because of the Manifold structure of \(\mans^3\).
The tangent space of the quaternion manifold can be interpreted as representing the rotation axis, scaled by half the rotation angle~\cite{zeestraten2018programming}.
Consequently, all dimensions interact with each other when mapping back and forth using the exponential and logarithmic maps.
To solve this problem, we apply an additional diagonal regularization after the frame transformation (parallel transport).
\figref{fig:fixedcov} shows the new covariance matrix and \figref{fig:fixedrec} the corrected reconstruction.
Other regularization schemes are possible, such as allowing correlations between the time and position dimensions.
Because positions are Euclidean, they do not suffer from the same problem introduced by parallel transport.
Note that such regularization also removes some of the desired correlations from the model, as can be seen from the missing rotation of the Gaussians in \figref{fig:fixedrec}.

\section{Experimental Details}\label{sec:exp_details}
\subsection{Baseline Hyperparameters}
For Diffusion Policy, we closely follow Chi~\etal{}~\cite{chi2023diffusionpolicy} in the selection of the hyperparameters.
We perform receding horizon control on absolute end-effector poses with an observation horizon of two and an action horizon of eight.
We also adopt the proposed cosine learning rate schedule.
However, when training on only five demonstrations, we reduce the number of warm-up steps from 500 to 50 to account for the reduced duration of the epochs.
For a full list of hyperparameters, please see refer to~\cite{chi2023diffusionpolicy} and to the accompanying code repository.
For setting up the LSTM, we follow von Hartz~\etal{}~\cite{vonhartz2023treachery} and utilize a two-layer LSTM to predict end-effector velocities.
To improve learning, we transform the task parameters into the gripper frame.
For both baselines, we perform linear normalization of both the policy inputs and outputs to improve gradient flow.

\subsection{Skill Segmentation and Parameter Selection}
We ablate the skill segmentation and task-parameter selection on three tasks: \texttt{SlideBlock}, \texttt{OpenDrawer}, and \texttt{PlaceCups}.
The first two are single-object tasks, but only \texttt{OpenDrawer} requires a precise grasping action, while \texttt{SlideBlock} does not.
\texttt{PlaceCups} not only adds multiple task-relevant objects but also additional clutter objects.
These three tasks are therefore a representative sample of all tasks.
We fit all models using identical hyperparameters.

\begin{figure*}
    \centering
  \includegraphics[width=\textwidth]{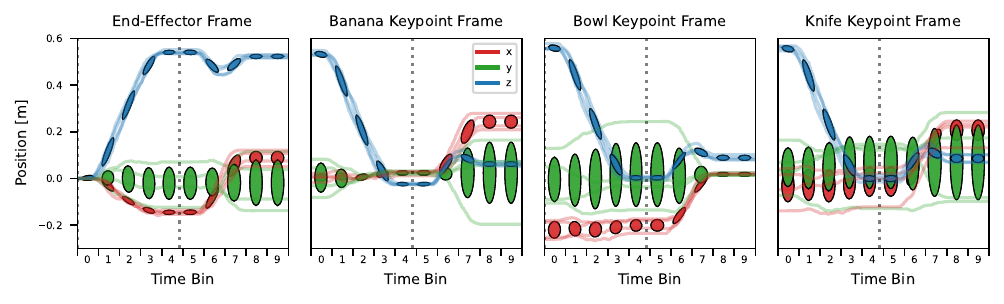}
  \caption{\rebuttal{Illustration of the parameter selection for the real world \texttt{PickAndPlace} task.
  The dotted line indicates the skill border between \emph{picking} and \emph{placing}.
  The temporal alignment of the per-frame models enables comparing their precision matrices over time, as describe in \secref{sec:cand_selec}.
  Of the shown four per-frame models, the End-Effector model has the highest relative precision across all three plotted dimensions in time bin \(0\), while the Banana model has the highest relative precision in bin \(4\).
  Consequently, both are selected for the \emph{picking} skill.
  For the \emph{placing} skill, the Banana's and the Bowl's frame are selected for analogous reason.
  The Knife is a clutter object whose marginal is relevant at no point in the task.}
  }\label{fig:frame_selection}
\end{figure*}

\begin{figure*}
    \centering
  \includegraphics{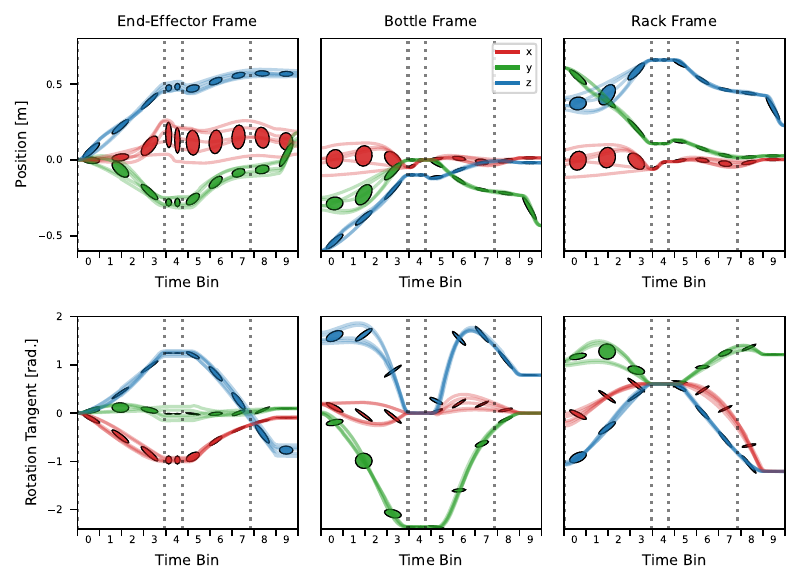}
  \caption{\rebuttal{Illustration of the parameter selection including the rotation component for the \texttt{StackWIne} task.
  The dotted lines indicate the skill borders.
  The rotation is given in tangent space and can be interpreted as the rotation axis scaled by \emph{half} the rotation angle (in radians).
  The relative precision of the per-frame marginals defined in \eqref{eq:frame_sel} uses the covariance matrix, which for the Riemmanian Gaussians is defined in the tangent space.
  Consequently, the frame selection takes into account the precision of the rotation as well.
  Note that the visual mismatch between the variance in the rotation trajectories and some of the plotted Gaussian components is due distortions from the tangent projection.
  As explained in \secref{sec:rieman_app}, the projection depends on the base at which the exponential or logarithmic map are applied.
  For plotting, we opted for using a consistent base, which leads to different levels of distortion at different points.
  }
  }\label{fig:frame_selection_wrot}
\end{figure*}

\subsection{State-Driven Policies}
\textit{HMM:} For this experiment, we slightly modify the Hidden Markov Model.
The resetting of the end-effector pose can lead to zero state probabilities across all states.
This can easily be seen from \eqref{eq:hmm}, which combines the current state distribution with the transition model and current observation probability.
The disturbance of the pose constitutes an unexpected state transition, leading to zero probabilities under the new observation.
Multiple possibilities exist to remedy this issue, such as allowing all state transitions with some small probability.
For simplicity, we allow the model to reset the prior \(\pi_k^{i_t}\) when all state probabilities become zero.

\textit{Disturbance:} We disturb the policies by freezing the end-effector for 90 timesteps in \texttt{SlideBlock} and 250 timesteps in \texttt{LiftBottle}.
The duration has been selected to correspond to the first contact point with the task object, such that the policies need to recover during a challenging part of the task.
We freeze the end-effector for \(n\) time steps, rather than resetting its pose after \(n\) steps, as resetting the pose directly can sometimes allow a state-based HMM that got stuck in a specific state to recover and we did not wish to mix these different effects.

\subsection{Real-World Policy Learning}
We use a single wrist-mounted Intel Realsense D435 camera to provide RGB-D observations.
We estimate the keypoint locations in 2D as described in \cite{vonhartz2023treachery, amir2021deep} and project them to 3D using the depth channel of the camera.
For each task, we manually select a set of five candidate task parameters on scene objects.
While automatic sampling from SAM-generated masks works well, manual sampling takes less than 30 seconds per task and allows for additional user control.
In particular, it allows us to ensure that the set of candidates represents a mix of task-relevant objects and clutter objects to challenge our task-parameter selection.
For example, in the \texttt{PickAndPlace} task, the policy needs to pick up a banana from a cutting board and place it in a bowl.
We select one candidate each on the banana, on the bowl, on the cutting board, on a cutter knife, and on the table.
Only the first two are relevant to the task, whereas the other three are distractors.
We also show an example of this procedure in the supplementary video.

\end{document}